\documentclass{article}
\usepackage{amsmath,graphicx,bm,stmaryrd,amsfonts,amsmath,algorithm,color,caption,subcaption,authblk,multirow,hyperref,fancyhdr}
\usepackage{tikz}
\usetikzlibrary{decorations.pathreplacing}
\hypersetup{hidelinks = true} 

\usepackage[small,compact]{titlesec}
\usepackage{xcolor}
\definecolor{simula}{HTML}{DF5429}

\twocolumn
\def\ninept{\def\baselinestretch{.95}\let\normalsize\small\normalsize}

\def\@maketitle{\newpage
 \null
 \vskip 3em \begin{center}
 {\large \bf \@title \par} \vskip 1.5em {\large \lineskip .5em
\begin{tabular}[t]{c}\@name \\ \@address
 \end{tabular}\par} \end{center}
 \par
 \vskip 1.5em}

\usepackage[noend]{algpseudocode}

\usepackage{amsmath}

\usepackage{amsfonts}

\usepackage{amssymb}

\usepackage{stmaryrd}

\usepackage[mathscr]{eucal}

\usepackage{amsbsy}

\usepackage{bm}

\usepackage{bbding}

\usepackage{array}

\usepackage{tabularx}

\usepackage{ctable}

\usepackage[neveradjust]{paralist}

\usepackage{fancyvrb}

\usepackage{caption}

\usepackage{placeins}

\usepackage{ifpdf}
\ifpdf
\DeclareGraphicsExtensions{.pdf,.png}
\else
\DeclareGraphicsExtensions{.eps}
\fi

\usepackage{algpseudocode}

\newcommand{\email}[1]{\href{mailto:#1}{\nolinkurl{#1}}}

\newcommand{\Sec}[1]{\hyperref[sec:#1]{\S\ref*{sec:#1}}} 
\newcommand{\Section}[1]{\hyperref[sec:#1]{Section~\ref*{sec:#1}}} 
\newcommand{\AppFull}[1]{\hyperref[sec:#1]{Appendix~\ref*{sec:#1}}} 
\newcommand{\Eqn}[1]{\hyperref[eq:#1]{(\ref*{eq:#1})}} 
\newcommand{\Fig}[1]{\hyperref[fig:#1]{Figure~\ref*{fig:#1}}} 
\newcommand{\Tab}[1]{\hyperref[tab:#1]{Table~\ref*{tab:#1}}} 
\newcommand{\Thm}[1]{\hyperref[thm:#1]{Theorem~\ref*{thm:#1}}} 
\newcommand{\Cor}[1]{\hyperref[cor:#1]{Corollary~\ref*{cor:#1}}} 
\newcommand{\Alg}[1]{\hyperref[alg:#1]{Algorithm~\ref*{alg:#1}}} 
\newcommand{\Def}[1]{\hyperref[def:#1]{Definition~\ref*{def:#1}}} 

\newcommand{\Real}{{\mathbb R}}

\newcommand{\Tra}{^{{\sf T}}} 
\newcommand{\Inv}{^{-1}} 
\newcommand{\V}[1]{{\bm{\mathbf{\MakeLowercase{#1}}}}} 
\newcommand{\Vhat}[1]{{\bm \hat{\mathbf{\MakeLowercase{#1}}}}} 

\newcommand{\M}[1]{{\bm{\mathbf{\MakeUppercase{#1}}}}} 
\newcommand{\MC}[2]{\V{#1}_{#2}} 
\newcommand{\MhatC}[2]{\Vhat{#1}_{#2}} 

\newcommand{\Kron}{\otimes} 
\newcommand{\Khat}{\odot} 
\newcommand{\Hada}{\ast} 

\newcommand{\T}[1]{\boldsymbol{\mathscr{\MakeUppercase{#1}}}} 

\newcommand{\norm}[1]{\left\lVert \, #1 \, \right\rVert}
\newcommand{\fnorm}[1]{\left\lVert \, #1 \, \right\rVert_{F}}

\let\oldthebibliography\thebibliography
\renewcommand\thebibliography[1]{%
  \oldthebibliography{#1}%
  \setlength{\itemsep}{0pt plus 0.3ex}%
}

\usepackage[numbers]{natbib}
\usepackage{mathtools}
\mathtoolsset{showonlyrefs}

\title{tPARAFAC2: Tracking evolving patterns in (incomplete) temporal data}

\author[1,2]{Christos Chatzis}
\author[1]{Carla Schenker}
\author[3]{Max Pfeffer}
\author[1]{Evrim Acar}
\affil[1]{Simula Metropolitan Center for Digital Engineering, Oslo, Norway}
\affil[2]{Faculty of Technology, Art and Design, OsloMet, Oslo, Norway}
\affil[3]{Institute for Numerical and Applied Mathematics, Georg-August-Universität Göttingen, Göttingen, Germany}
\date{}
\begin{document}

\maketitle

\begin{abstract}
Tensor factorizations have been widely used for the task of uncovering patterns in various domains. Often, the input is time-evolving, shifting the goal to tracking the evolution of the underlying patterns instead. To adapt to this more complex setting, existing methods incorporate temporal regularization but they either have overly constrained structural requirements or lack uniqueness which is crucial for interpretation. In this paper, in order to capture the underlying evolving patterns, we introduce t(emporal)PARAFAC2, which utilizes temporal smoothness regularization on the evolving factors. Previously, Alternating Optimization (AO) and Alternating Direction Method of Multipliers (ADMM)-based algorithmic approach has been introduced to fit the PARAFAC2 model to fully observed data. In this paper, we extend this algorithmic framework to the case of partially observed data and use it to fit the tPARAFAC2 model to complete and incomplete datasets with the goal of revealing evolving patterns. Our numerical experiments on simulated datasets demonstrate that tPARAFAC2 can extract the underlying evolving patterns more accurately compared to the state-of-the-art in the presence of high amounts of noise and missing data. Using two real datasets, we also demonstrate the effectiveness of the algorithmic approach in terms of handling missing data and tPARAFAC2 model in terms of revealing evolving patterns. The paper provides an extensive comparison of different approaches for handling missing data within the proposed framework, and discusses both the advantages and limitations of tPARAFAC2 model.
\end{abstract}

\section{Introduction}\label{sec1}

Temporal datasets capture the evolution of an event or a system as a sequence of time-stamped observations. Uncovering latent patterns in such temporal datasets and the evolution of those patterns over time is crucial across various domains to gain insights about complex systems. For example, in neuroscience, capturing spatial networks of brain connectivity as well as their evolution over time holds the promise to improve our understanding of brain function \cite{AcRoHo22}. Other examples are from social network analysis, where evolving patterns may reveal changes in communities of users, or dynamic topic modeling which aims to detect the temporal evolution of topics, for instance, within a large collection of documents \cite{Ahn2021, DiRu19}.

\begin{figure}[t]
    \begin{minipage}[b]{1.0\linewidth}
      \centering      \centerline{\includegraphics[width=\textwidth]{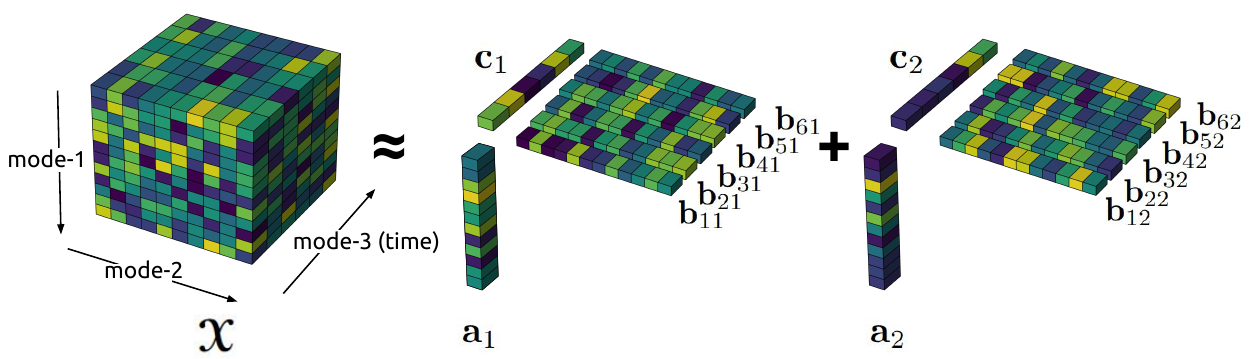}}
      \caption{Illustration of a 2-component PARAFAC2 model of a third-order tensor, where mode 2 (with $\V{b}_{kr}$, for $k=1,...,6$ and $r=1,2$) is the evolving mode.}
    \label{fig:parafac2}
    \end{minipage}
\end{figure}

Temporal data frequently takes the form of a multidimensional array (i.e., a tensor) and tensor factorizations have been an effective tool for their analysis. The rationale of these unsupervised methods is to decompose the input into compact and interpretable factors that reflect the main trends present. For example, \cite{enron} applied the CANDECOMP/PARAFAC (CP) decomposition \cite{parafac,parafac_} on Enron email data (arranged as a third-order tensor with modes: authors, words, and time) and extracted interpretable factors directly connected to the company's collapse. \cite{parafac-qos} incorporated non-negativity constraints, allowing CP to improve prediction accuracy on missing entries on QoS (Quality-of-Service) data relevant to online services. Nevertheless, these methods do not incorporate intrinsic properties of the temporal dimension of the input. Thus, extensions of CP tailored for temporal data have been proposed \cite{trmf,tatd,seekanddestroy}. \cite{trmf} enhanced the CP decomposition with autoregressive regularization on the temporal and dimension, which allowed for more accurate forecasting. Time-aware Tensor Decomposition (TATD) \cite{tatd} assumes smooth changes across data of consecutive time points and simultaneously imposes sparsity regularization with strength that varies over time. SeekAndDestroy \cite{seekanddestroy} updates a CP decomposition in an online fashion as new data is received, and keeps track of disappearing and emerging concepts. In all of these methods, nonetheless, the concepts uncovered by the factors must adhere to the CP structure, which means they are only allowed to change by a scalar multiplicator over time. This is a limitation when the task at hand is to discover \textit{evolving} patterns.

In terms of capturing time-evolving patterns, other proposed methods in the literature follow the less structurally-constrained Collective Matrix Factorization (CMF) paradigm \cite{cmf}, while incorporating additional temporal regularization \cite{tmf,chimera,smf,ssmf}. \cite{tmf} proposed Temporal Matrix Factorization (TMF), which factorizes the input using a set of time-index-dependent factors. The method proposed by  \cite{chimera} receives additional contextual inputs and factorizes the input assuming smooth changes over time. Under the same assumption of small changes across consecutive time points, \cite{smf} proposed SMF (Seasonal Matrix Factorization), an online method that involves updating the factors through small gradient descent steps to facilitate smooth factor evolution while also addressing the seasonality of uncovered patterns. This work has later been extended to accept multiple seasonal `regimes' \cite{ssmf}. While CMF-based approaches provide more expressive factors that may reflect the underlying evolving patterns in greater detail, they generally do not possess uniqueness guarantees for their solutions (or their uniqueness properties are not well-studied), which are crucial for interpretability \cite{reproducibility}.

In the middle ground between CP and CMF-based approaches lies the PARAFAC2 factorization \cite{parafac2}, a technique that has shown its effectiveness across various fields such as in chemometrics in terms of analyzing measurements of samples with unaligned profiles \cite{bro1999parafac2}, in neuroscience by allowing for subject-specific temporal profiles \cite{MaCh17} and task-specific spatial maps \cite{multitaskfmri}, and in electronic health record (EHR) data analysis \cite{PePa19}. Apart from having uniqueness properties (under certain conditions, \cite{parafac2_direct}), PARAFAC2 offers the flexibility of allowing the factors to vary along one specific mode (evolving mode) in all these applications. Several adaptations of PARAFAC2 for temporal data have been proposed, each incorporating different temporal regularization techniques \cite{logpar,copa,repair,atom,tedpar}. LogPar \cite{logpar} regularizes the factor matrix in the time mode by assuming that columns of the factor matrix slowly change, and uses an exponential decay-based weight. COPA \cite{copa} and REPAIR \cite{repair} utilize M-spline-based temporal smoothness on the factor matrix extracted from the time mode. ATOM \cite{atom} ensures the smoothness of the components in the time mode by penalizing the difference between temporally consecutive factors.
TedPar \cite{tedpar} is a PARAFAC2-based framework for EHR data that utilizes temporal smoothness to monitor smooth phenotypic changes. However, all these studies focus on the regularization of time mode factors.

In this paper, we use the PARAFAC2 model to extract evolving patterns in time (as in Figure \ref{fig:parafac2}, where mode 3 is time and the evolving mode (mode 2) may correspond to, \textit{e.g.}, voxels to capture evolving spatial maps,  words to capture evolving topics) and consider temporal regularization of such time-evolving patterns. In other words, we force the structure of the patterns to change smoothly over time and not their strength. Our contributions can be summarized as follows:
\begin{itemize}
    \item We introduce the t(emporal)PARAFAC2 model and an Alternating Direction Method of Multipliers (AO-ADMM)-based algorithmic approach to fit the model, 
    \item We consider two different ways of handling missing data, \textit{i.e.}, an Expectation Maximization (EM)-based approach and one employing Row-Wise (RW) updates, when fitting a regularized PARAFAC2 model using the AO-ADMM framework. As regularized PARAFAC2 encompasses tPARAFAC2, this enables fitting tPARAFAC2 to incomplete datasets, 
    \item Using extensive numerical experiments on  synthetic data with slowly changing patterns with varying levels of noise and amounts of missing data, we demonstrate that tPARAFAC2 can reveal the underlying evolving patterns more accurately compared to alternative approaches, in the case of high levels of noise and missing data,
    \item We demonstrate that while the EM-based approach and the RW updates are equally accurate, the EM-based approach is computationally more efficient,
    \item We use the proposed methods on two real datasets. In a chemometrics application, we demonstrate that the extension of the AO-ADMM algorithm for regularized PARAFAC2 to incomplete data can recover the underlying patterns accurately in the presence of missing data. In a metabolomics application, we use PARAFAC2 and tPARAFAC2 to capture evolving metabolite patterns from a dynamic metabolomics dataset, and show the effectiveness of tPARAFAC2 in the case of high amounts of missing data.
\end{itemize}

This paper is an extension of our preliminary study \cite{tparafac2} which introduced t(emporal)PARAFAC2 and used an AO-ADMM-based algorithmic approach to fit the model. Our preliminary results demonstrated the promise of tPARAFAC2 in noisy and low-signal settings considering patterns following various types of change in time  \cite{tparafac2}. Here, we focus on slowly changing patterns and study the performance of the tPARAFAC2 model at different levels of noise and with (structured/random) missing data. We extend the algorithmic framework to incomplete data considering two different approaches to handle missing entries. The proposed approaches have also been used in two real applications, one from chemometrics focusing on modeling incomplete data and the other from metabolomics focusing on extracting evolving patterns.
 
The rest of the paper is organized as follows: Section \ref{sec:Preliminaries} describes the notation used in this work, and briefly discusses the PARAFAC2 model as well as the algorithmic approaches for fitting a PARAFAC2 model. Section \ref{sec:proposed} introduces the proposed tPARAFAC2 model and the AO-ADMM algorithm to fit the model. In subsection \ref{subsec:tparafac2-missing}, we introduce two approaches to extend the AO-ADMM algorithm for regularized PARAFAC2 to incomplete datasets and discuss them with respect to the state-of-the-art while section \ref{sec:experiments} contains numerical experiments. We conclude with a summary of the findings, limitations of the proposed work and potential future research directions in section \ref{conclusion}.
\section{Preliminaries}
\label{sec:Preliminaries}
\subsection{Notation}
\label{sec:notation}

In this paper, we adopt the tensor notation of \cite{KoBa09}. The order of a tensor indicates the number of modes; in other words, the number of indices needed to specify an entry. In this regard, a vector is a tensor of order one and a matrix is a tensor of order two. Tensors with order of at least three are referred to as higher-order tensors. We use bold lowercase letters to denote vectors (e.g. {\small$\V{x}$}, 
 {\small$\V{y}$}) and bold capital letters for matrices (e.g. {\small$\M{x}$}, {\small$\M{y}$}). For higher-order tensors, we use bold uppercase Euler script letters (e.g. {\small$\T{x}$}, {\small$\T{y}$}). Table \ref{tab:tensor_products} shows our notation for various tensor-related mathematical operations.

\begin{table}[ht]
\centering
\renewcommand{\arraystretch}{2.0}
\captionsetup{justification=centering, singlelinecheck=false, width=\textwidth}
\caption{Notation used for relevant mathematical operations.}
\begin{tabular}{|>{\centering\arraybackslash}p{2.25cm}|>{\centering\arraybackslash}p{4.4cm}|}
\hline
\textbf{Symbol} & \textbf{Operation} \\
\hline
\hline
$\circ$ & Outer product \\
\hline
$\Kron$ & Kronecker product \\
\hline
$\Khat$ & Khatri-Rao product \\
\hline
$\Hada$ & Hadamard product \\
\hline
$\M{X}\Tra$ & Matrix transpose \\
\hline
$\M{X}\Inv$ & Matrix inverse \\
\hline
$\fnorm{\cdot}$ & Frobenius norm \\
\hline
$diag(\M{A})$ & diag operation\footnotemark[1] \\
\hline
\end{tabular}
\footnotetext[1]{https://www.mathworks.com/help/matlab/ref/diag.html}
\label{tab:tensor_products}
\end{table}

It is useful in practice to refer to specific segments of a tensor. The higher-order equivalent of matrix rows and columns is referred to as fibers. We can obtain mode-{\small$n$} fibers of a tensor by fixing the indices on all modes except the {\small$n$}-th. In the same manner, we can obtain (matrix) slices of a tensor by fixing all indices except two. To exemplify these concepts, consider a third-order tensor {\small$\T{X} \in \Real^{I \times J \times K}$}: We can obtain any mode-{\small$1$} fiber of {\small$\T{X}$} by fixing {\small$j$} and {\small$k$} in {\small$\T{X}{(:,j,k)}\in\Real^{I}$} and any ``frontal" slice by fixing {\small$k$} in  {\small$\T{X}{(:,:,k)}\in\Real^{I \times J}$}. Specifically for frontal slices, we use the shorter notation {\small$\M{X}_{k}$}.

\subsection{The PARAFAC2 factorization} \label{sec:parafac2}

Originally proposed by Harshman \cite{parafac2}, PARAFAC2 models each frontal slice of a third order tensor {\small$ \T{X} \in \Real^{I \times J \times K}$} as a product of three factors:{\small
\begin{equation} \label{parafac2-slicewise}
\begin{split}
    \M{X}_k \approx \M{A} \M{D}_k \M{B}_k\Tra \\
    {\{\M{B}_k}\}_{k=1}^{K} \in \mathscr{P}
\end{split}\end{equation}}where {\small$\M{A}\in\Real^{I \times R}$}, {\small$\M{D}_k\in\Real^{R \times R}$} is a diagonal matrix and {\small$\M{B}_k\in\Real^{J \times R} \; \forall \; k=1,\dots,K$}. $R$ denotes the number of components. It is convenient to concatenate the diagonals of the {\small$\{\M{D}_k\}_{k=1}^{K}$} as rows in a single factor matrix {\small$\M{C}\in\Real^{K \times R} $}. Figure \ref{fig:parafac2} is an illustration of a 2-component PARAFAC2 model (\textit{i.e.} \small $R=2$) and six frontal slices (\textit{i.e.} {\small$K=6$}). In this specific case:
\begin{align*}
    \M{A} &= \begin{bmatrix}
        | & | \\
        \V{a_1} & \V{a_2} \\
        | & | \\
    \end{bmatrix},
    &
    \M{C} &= \begin{bmatrix}
        | & | \\
        \V{c_1} & \V{c_2} \\
        | & | \\
    \end{bmatrix}
\end{align*}

\begin{equation*}
    \text{and} \ \ \M{B}_k = \begin{bmatrix}
        | & | \\
        \V{b_{k1}} & \V{b_{k2}} \\
        | & | \\
    \end{bmatrix} \quad \forall k = 1, \dots, 6 \text{.}
\end{equation*}

The second line of Equation \eqref{parafac2-slicewise} denotes the constant cross-product constraint of PARAFAC2:
\begin{equation} \label{parafac2-constraint}
    \mathscr{P} = \left\{ \{\M{B}_k\}_{k=1}^{K} \mid \M{B}\Tra_{k_{1}}\M{B}_{k_{1}} = \M{B}\Tra_{k_{2}}\M{B}_{k_{2}} \ \ \forall k_{1},\! k_{2} \in \{1, 2, ... ,K\}  \right\} .
\end{equation}We refer to this constraint as the PARAFAC2 constraint. In essence, this constraint requires all evolving factors to share the same Gram matrix and the inner products between components (i.e. columns of {\small $\M{B}_k$}) -or their correlations, if normalized- to remain constant across {\small $k$} \citep{parafac2_direct}.

PARAFAC2 is considered to have an ``essentially'' unique solution up to permutation and scaling under certain conditions \citep{parafac2_direct}. To better understand the ambiguities present in the solution, consider the formulation of PARAFAC2 for a specific slice as follows:
{\small
\begin{equation} \label{parafac2-slicewise-vector}
    \M{X}_k \approx \sum_{r=1}^{R} \V{d}_{kr} ( \V{a}_r \circ \V{b}_{kr})
\end{equation}}where {\small$\V{d}_{kr} = \M{D}_{k}(r,r)$}, {\small$\V{a}_r = \M{A}{(:,r)}$} and {\small$\V{b}_{kr} = \M{B}_{k}(:,r)$}. Reordering the components of the summation does not impact the quality of the solution, constituting a permutation ambiguity. Moreover, if any of {\small$\V{d}_{kr}$}, {\small $\V{a}_r$} and {\small$\V{b}_{kr}$} is scaled by {\small$\gamma$} and the other two are scaled {simultaneously by factors with product {\small$\frac{1}{\gamma}$} (the PARAFAC2 constraint should be satisfied at all times), the model estimate of the slice is consistent and the solution is considered equivalent (scaling ambiguity). While such ambiguities do not interfere with the interpretation of the captured factors, there is an additional challenge due to the sign ambiguity in PARAFAC2 since 
{\small$\V{d}_{kr}$} can arbitrarily flip signs together with {\small$\V{b}_{kr}$} \citep{parafac2}. One possible solution to fix the sign ambiguity is to impose non-negativity constraints on {\small$\M{C}$} \citep{parafac2, parafac2_direct}.

\subsection{Algorithms for fitting PARAFAC2}

\subsubsection{Alternating Least Squares (ALS)}
Much work in the literature has focused on finding an efficient algorithm for fitting the PARAFAC2 model. \cite{parafac2_direct} proposed the `direct fitting' method, which parametrizes each {\small$\M{b}_k$} as {\small$\M{b}_k = \M{P}_k \M{B}$}, where each {\small$\M{P}_k$} is constrained to have orthonormal columns. Fitting the PARAFAC2 model is then formulated as the following optimization problem:
{\small
\begin{equation} \label{parafac2-direct}
\begin{split} 
     \min_{\M{A},\{\M{P}_k\}_{k=1}^K,\M{B},\M{C}} & \quad \left\{\sum_{k=1}^{K} \fnorm{\M{X}_k \!-\! \M{A} \M{D}_k (\M{P}_k \M{B})\Tra }^2 \right\} \\
     \text{subject to} \quad & \quad  \M{P}_k\Tra \M{P}_k = \M{I} \quad  \forall \; k=1,\dots,K .
\end{split}\end{equation}}This parametrization encapsulates the PARAFAC2 constraint as {\small$\M{B}_k\Tra \M{B}_k = \M{B}\Tra  \M{P}_k\Tra \M{P}_k \M{B} = \M{B}\Tra \M{B}$}, which is constant across all values of $k$. Fixing all factors except {\small$\{\M{P}_k\}_{k=1}^{K}$} results in orthogonal Procrustes problems and Singular Value Decomposition (SVD) is used to solve them. Then, each frontal slice is multiplied by $\M{P}_k$ and problem \eqref{parafac2-direct} is minimized with respect to {\small$\M{A}$}, {\small$\M{B}$} and {\small$\M{C}$} by alternatingly solving the respective least squares problems until convergence. This approach is referred to as the PARAFAC2 ALS algorithm.

Imposing additional constraints is often beneficial, for example, to enhance interpretability, incorporate domain knowledge, or improve robustness to noise. While it is possible to impose certain constraints on factor matrices {\small$\M{A}$} and {\small$\M{C}$} using an ALS-based algorithm when solving problem \eqref{parafac2-direct}, it is difficult to do so for {\small$\{ \M{B}_k \}_{k=1}^{K}$} factor matrices. Recently, the `flexible coupling' approach, which formulates the PARAFAC2 constraint as an additional penalty term, was introduced to incorporate non-negativity constraints in all modes \citep{parafac2-flexiblecoupling}. In theory, some other constraints could also be
incorporated by employing specialized constrained least-squares
solvers. However, as described by \cite{parafac2-flexiblecoupling}, the method suffers from the time-consuming tuning of the penalty parameter.

\subsubsection{AO-ADMM} A more flexible algorithmic approach for fitting the PARAFAC2 model is the AO-ADMM-based algorithmic approach, which facilitates the use of a variety of constraints on all modes \citep{parafac2-aoadmm}. The AO-ADMM-based approach formulates the regularized PARAFAC2 problem as:
{\small
\begin{equation} \label{parafac2-aoadmm}
\begin{split} 
     \min_{\M{A},\{\M{B}_k\}_{k=1}^K,\M{C}} \quad & \Bigg\{ \sum_{k=1}^{K} \fnorm{ \M{X}_k \!-\! \M{A} \M{D}_k \M{B}_k^T}^2 + g_{\M{A}}(\M{A})  \\
     & + g_{\M{B}}(\{\M{B}_k\}_{k=1}^{K}) + g_{\M{D}}(\{\M{D}_k\}_{k=1}^{K}) \Bigg\} \\
    \text{subject to} \; \; \quad & \quad \{\M{B}_k\}_{k=1}^{K} \in  \mathscr{P} \quad  \forall \; k=1,\dots,K 
\end{split}
\end{equation}}where {\small$g_{\M{A}}$}, {\small $g_{\M{B}}$} and {\small $g_{\M{D}}$} denote regularization penalties imposed on respective factors. In principle, after the augmented Lagrangian is formed, the algorithm alternates between solving the subproblems for each of the factor matrices using ADMM. To solve the non-convex subproblems of the evolving factors {\small$\{\M{B}_k\}_{k=1}^{K}$}, an effective iterative scheme is proposed to project the evolving factors onto {\small$\mathscr{P}$} \citep{parafac2-aoadmm}. This procedure is repeated until changes in the solution are sufficiently small, which is indicated by small relative or absolute change in the objective function of Equation \eqref{parafac2-aoadmm}, and as long as the constraints are satisfied adequately (i.e. small `feasibility' gaps):
{\small
\begin{align*}
    \lVert f^{(n+1)} - f^{(n)} \rVert  < \epsilon_{abs} \ \ &\textit{OR} \ \ \frac{\lVert f^{(n+1)} - f^{(n)} \rVert}{ \lVert f^{(n)} \rVert } < \epsilon_{rel} \\
    \textit{AND} \  \frac{\norm{ \M{M} - \M{Z}_{\M{M}} }}{\norm{\M{M}}} &< \epsilon_{feasibility}
\end{align*}
}where {\small $f^{(n+1)}$} and {\small $f^{(n)}$} denote the function value at iteration {\small $n+1$} and {\small $n$}, respectively, and for each factor matrix {\small $\M{M}$}, {\small $\M{Z}_{\M{M}}$} denotes the respective auxiliary variable. The last condition should hold for all factor matrices that involve regularization. The tolerances {\small $\epsilon_{abs}$}, {\small $\epsilon_{rel}$} and {\small $\epsilon_{feasibility}$} are user-defined. More information about the exit conditions can be found in the supplementary material.

\section{Proposed methodology}
\label{sec:proposed}

\subsection{tPARAFAC2} \label{subsec:tparafac2}
When PARAFAC2 is used to analyze time-evolving data with frontal slices {\small$\M{X}_k$} changing in time, {\small$k = 1, ...,K$}, {\small$\{\M{B}_k\}_{k=1}^{K}$} factor matrices can capture the structural changes of the patterns in time while {\small$\{\M{D}_k\}_{k=1}^{K}$} capture the strength over time. Combining information from both sets of factors allows for a thorough understanding of the underlying evolving patterns. For example, this key property of PARAFAC2 has been previously used to uncover evolving spatial brain activation maps (i.e., spatial dynamics) from fMRI data \citep{parafac2_evolution, AcRoHo22}.

However, without additional constraints, PARAFAC2 is not time-aware, i.e., the model is unable to take into account the sequential nature of the temporal dimension. One can confirm this by applying the factorization on two reordered versions of a dataset: the factorizations will be respectively reordered. While mathematically acceptable, the two versions could potentially describe two different event sequences.

To tackle this problem, we introduce t(emporal)PARAFAC2, which is formulated as \citep{tparafac2}:
{\small
\begin{equation} \label{tparafac2-abstract}
\begin{split} 
     \min_{\M{A},\{\M{B}_k\}_{k=1}^K,\M{C}} \quad & \Bigg\{ \sum_{k=1}^{K} \fnorm{\M{X}_k \!-\! \M{A} \M{D}_k \M{B}_k^T}^2 + g_{\M{A}}(\M{A}) \\ 
     & + \lambda_\M{B} \sum_{k=2}^K\fnorm{ \! \M{B}_k \!-\! \M{B}_{k-1} \! }^{2} + g_{\M{D}}(\{\M{D}_k\}_{k=1}^{K}) \Bigg\} \\
    \text{subject to} \; \; \quad & \quad \{\M{B}_k\}_{k=1}^{K} \in  \mathscr{P} \quad  \forall \; k=1,\dots,K \ .
\end{split}
\end{equation}}}Equation \eqref{tparafac2-abstract} is essentially the regularized PARAFAC2 AO-ADMM problem (Equation \eqref{parafac2-aoadmm}) with {\small $g_{\M{B}}(\{\M{B}_k\}_{k=1}^{K}) = \lambda_B \sum_{k=2}^K\fnorm{ \M{B}_k \!-\! \M{B}_{k-1} }^{2}$}. This formulation assumes that the data exhibits fine temporal granularity or that the underlying patterns evolve slowly over time - similar to the temporal regularization previously used in CMF \citep{chimera}. Therefore, tPARAFAC2 regularizes {\small$\M{B}_k$} factor matrices that correspond to consecutive time slices to be similar, with the regularization strength controlled by {\small$\lambda_\M{B}$}. Due to the scaling ambiguity, it is crucial to have norm-based regularization in the other modes \citep{parafac2-aoadmm}. Furthermore, to overcome the sign ambiguity of PARAFAC2, we impose non-negativity on the factors of the third mode. Therefore, in the optimization problem \eqref{tparafac2-abstract} we set
{\small
\begin{align*}
    g_{\M{A}}(\M{A}) &= \lambda \fnorm{\M{A}}^{2} \\
    g_{\M{D}}(\{\M{D}_k\}_{k=1}^{K}) &= \sum_{k=1}^{K} \bigl( \lambda \fnorm{\M{D}_k}^2 + \iota_{\mathbb{R}_{+}}(\M{D}_k)  \bigr)
\end{align*}}where hyperparameter {\small $\lambda$} control the strength of the ridge penalties and {\small $\iota_{\mathbb{R}_{+}}(\cdot)$} is the indicator function for the non-negative orthant.

\subsection{Optimization} \label{subsec:tparafac2-opt}

\begin{algorithm}[t]
\caption{tPARAFAC2: ADMM mode-2 ($\M{B}_k$) updates}\label{admm-B}
 \textbf{Input:} Data Input {\small{$ \T{X},\M{A},\{{{\M{B}_k},\M{Y}_{\M{B}_k}},\M{Z}_{\M{B}_k},\M{\mu}_{\M{Z}_{\M{B}_k}},\M{\mu}_{\M{\Delta}_{\M{B}_k}}\}_{k=1}^K,\M{C}$}.} \\
\textbf{Output:} {\small ${\{{{\M{B}_k},\M{Y}_{\M{B}_k}},\M{Z}_{\M{B}_k}},\M{\mu}_{\M{Z}_{\M{B}_k}},\M{\mu}_{\M{\Delta}_{\M{B}_k}}\}_{k=1}^K$.}
\begin{algorithmic}[1]
\State \textbf{while} \textit{stopping conditions are not met}
\State \hspace{0.25cm} \textbf{for} {\small $k \gets 1$ \normalsize \textbf{to} {\small $K$ }}\textbf{do}
\State  \hspace{0.55cm} {\small$\M{B}_k \gets \arg\min_{{\M{B}_k}} \left\{ \mathcal{L}  \right\}$} \algorithmiccomment{Equation \eqref{b-update}}
\State \hspace{0.25cm} \textbf{end}
{\small \State \hspace{0.25cm} $ {\{\M{Z}_{\M{B}_K}}\}_{k=1}^K \!\gets\! \text{\emph{Tridiagonal solve}}\! \left(\! \lambda_{\M{B}},\!\{\M{B}_k,\rho_{\M{B}_k}\!,\!\M{\mu}_{\M{Z}_{\M{B}_k}}\}_{k=1}^K \right) $}
{\small \State \hspace{0.25cm} $ {\{\M{Y}_{\M{B}_K}}\}_{k=1}^K \!\gets\! \text{\cite[Algorithm 5]{parafac2-aoadmm}}$}
\State \hspace{0.25cm} \textbf{for} {\small $k \gets 1$ \normalsize \textbf{to} {\small $K$ }}\textbf{do}
\State {\small \hspace{0.55cm} $\M{\mu}_{\M{Z}_{\M{B}_k}} \gets \M{B}_k - \M{Z}_{\M{B}_k} + \M{\mu}_{\M{Z}_{\M{B}_k}} $}
\State {\small \hspace{0.55cm} $\M{\mu}_{\M{\Delta}_{\M{B}_k}} \gets \M{B}_k - \M{Y}_{\M{B}_k} + \M{\mu}_{\M{\Delta}_{\M{B}_k}} $}
\State \hspace{0.25cm} \textbf{end}
\State \textbf{end}
\end{algorithmic}
\end{algorithm}

\begin{algorithm}[t]
\caption{tPARAFAC2 AO-ADMM} \label{tparafac2-aoadmm}
\textbf{Input:} Data input {\small$\T{X}$}. \\
\textbf{Output:} PARAFAC2 factors {\small$\M{A},\{\M{B}_k\}_{k=1}^{K},\{\M{D}_k\}_{k=1}^{K}$}.
\begin{algorithmic}[1]
\State \textit{Initialize} {\small$\M{A}$}
\State \textit{Initialize} {\small$\{\M{B}_k,\!\M{Y}_{\M{B}_k},\M{Z}_{\M{B}_k},\M{\mu}_{\M{Z}_{\M{B}_k}},\M{\mu}_{\M{\Delta}_{\M{B}_k}}\}_{k=1}^{K}$}
\State \textit{Initialize} {\small$\{\M{D}_k,\M{Z}_{\M{D}_k},\M{\mu}_{\M{D}_k}\}_{k=1}^{K} $}
\State \textbf{while} \textit{outer stopping conditions are not met}
\State  \hspace{0.25cm} {\small$ \{{{\M{B}_k},\M{Y}_{\M{B}_k}},\M{Z}_{\M{B}_k},\M{\mu}_{\M{Z}_{\M{B}_k}},\M{\mu}_{\M{\Delta}_{\M{B}_k}}\}_{k=1}^K \gets \text{Algorithm \ref{admm-B}}$}
\State \hspace{0.25cm} {\small$\{{{\M{D}_k},\M{Z}_{\M{D}_k}},\M{\mu}_{\M{D}_k}\}_{k=1}^K \gets \text{\cite[Algorithm 7]{parafac2-aoadmm}}$}
\State \hspace{0.25cm} {\small$\M{A} \gets \arg \min_{\M{A}} \{ \mathcal{L} \} $} \algorithmiccomment{Equation \eqref{a-update}}
\State \textbf{end} 
\end{algorithmic}
\end{algorithm}

We solve the optimization problem \eqref{tparafac2-abstract} using the PARAFAC2 AO-ADMM framework \citep{parafac2-aoadmm}. The augmented Lagrangian of the full optimization problem is given by:
{\small \begin{align} 
    &\mathcal{L} =\sum_{k=1}^{K}\lVert \M{X}_k - \M{A}\M{D}_k\M{B}_k\Tra\rVert^2_{F} + \lambda \lVert \M{A} \rVert^{2}_F + \lambda \sum_{k=1}^K \lVert \M{D}_k \rVert^{2}_F \notag  \\
    &+ \lambda_\M{B} \sum_{k=2}^K\lVert \M{Z}_{\M{B}_k}-\M{Z}_{\M{B}_{k-1}} \rVert^2_F + \sum_{k=1}^K \frac{\rho_{\M{B}_k}}{2}\lVert \M{B}_k -  \M{Z}_{\M{B}_k} + \M{\mu}_{\M{Z}_{\M{B}_k}} \rVert^2_{F} \notag \\
    &+ \iota_{\mathscr{P}}(\{\M{Y}_{\M{B}_k}\}_{k=1}^K) + \sum_{k=1}^K \frac{\rho_{\M{B}_k}}{2}\lVert \M{B}_k  - \M{Y}_{\M{B}_k} + \M{\mu}_{\M{\Delta}_{\M{B}_k}} \rVert^2_{F} \notag \\
    &+ \sum_{k=1}^K\iota_{\mathbb{R}_{+}}(\M{Z}_{\M{D}_{k}})  + \sum_{k=1}^K \frac{\rho_{\M{D}_{k}}}{2}\lVert  \M{D}_k  - \M{Z}_{\M{D}_{k}}  + \M{\mu}_{\M{D}_{k}} \rVert^2_{F}
    \label{eq:tparafac2-lagrangian}
\end{align}}where we have introduced auxiliary variables {\small $\{ \M{Z}_{\M{B}_k},\M{Y}_{\M{B}_k},\M{Z}_{\M{D}_k} \}_{k=1}^{K}$}, dual variables {\small $\{ \M{\mu}_{\M{Z}_{\M{B}_k}},\M{\mu}_{\M{\Delta}_{\M{B}_k}},\M{\mu}_{_{\M{D}_k}} \}_{k=1}^{K}$}, step-sizes {\small $\{\rho_{\M{B}_k},\rho_{\M{D}_k}\}_{k=1}^{K}$} and
{\small
\begin{equation*}
    \iota_{\mathscr{P}}(\{\M{Y}_{\M{B}_k}\}_{k=1}^{K}) = \begin{cases}
        0 \ \  \ \text{if } \{\M{Y}_{\M{B}_k}\}_{k=1}^{K}\in\mathscr{P}  \\
        \infty \  \text{otherwise}
    \end{cases}
\end{equation*}}is the indicator function for the sets of matrices that satisfy the PARAFAC2 constraint.

We use AO-based optimization to solve for each of the factor matrices independently. Fixing all variables and solving for {\small \M{A}} yields the following update rule:
{\small
\begin{equation} \label{a-update}
     \frac{\partial\mathcal{L}}{\partial\M{A}} = 0 \Leftrightarrow \M{A}^{*} \gets \bigl( \sum_{k=1}^{K} \M{X}_k \M{B}_k \M{D}_k \bigr) \bigl(  \sum_{k=1}^{K} \M{D}_k \M{B}_k\Tra \M{B}_k \M{D}_k + \lambda \M{I} \bigr)^{-1}
\end{equation}}For the evolving factors, we use ADMM. The update rule for each {\small$\M{B}_k$} can be formulated similarly:
{\small
\begin{equation} \label{b-update}
     \frac{\partial\mathcal{L}}{\partial\M{B}_k} = 0 \Leftrightarrow \M{B}_k^{*} \gets \bigl( \M{X}_k\Tra \M{A} \M{D}_k + \frac{\rho_{\M{B}_k}}{2} \M{M} \bigr) \bigl( \M{D}_k \M{A}\Tra \M{A} \M{D}_k + \rho_{\M{B}_k} \M{I} \bigr)^{-1}
\end{equation}}where {\small $\M{M}=\M{Z}_{\M{B}_k} - \M{\mu}_{\M{Z}_{\M{B}_k}} +\M{Y}_{\M{B}_k} - \M{\mu}_{{\M{\Delta}_{\M{B}_k}}} $}. The problem of minimizing for each {\small $\M{Z}_{\M{B}_k}$} is not separable for different values of {\small$k$}. Thus, we have to differentiate {\small $\mathcal{L}$} with respect to each {\small $\M{Z}_{\M{B}_k}$}:
{\small
\begin{align*} \label{zbk-update}
    \frac{\partial\mathcal{L}}{\partial \M{Z}_{\M{B}_1}} &= 0 \Leftrightarrow \bigl( 2\lambda_\M{B} + \rho_{\M{B}_1} \bigr) \M{Z}_{\M{B}_1} - 2\lambda_\M{B} \M{Z}_{\M{B}_2} = \rho_{\M{B}_1} \bigl( \M{B}_1 + \M{\mu}_{\M{Z}_{\M{B}_1}} \bigr) \\
    \frac{\partial\mathcal{L}}{\partial \M{Z}_{\M{B}_k}} &= 0 \Leftrightarrow \bigl( 4\lambda_\M{B} + \rho_{\M{B}_k} \bigr) \M{Z}_{\M{B}_k} - 2\lambda_\M{B} \bigl( \M{Z}_{\M{B}_{k-1}} + \M{Z}_{\M{B}_{k+1}} \bigr) \\
    &\quad = \rho_{\M{B}_k} \bigl( \M{B}_k + \M{\mu}_{\M{Z}_{\M{B}_k}} \bigr) \ \forall \ k = 2,\ldots,K-1 \\
    \frac{\partial\mathcal{L}}{\partial \M{Z}_{\M{B}_K}} &= 0 \Leftrightarrow \! \bigl( 2\lambda_\M{B} \!+\! \rho_{\M{B}_K} \bigr) \M{Z}_{\M{B}_K} \!\!-\! 2\lambda_\M{B} \M{Z}_{\M{B}_{K\!-\!1}} \!\!=\! \rho_{\M{B}_K} \bigl( \M{B}_K \!+\! \M{\mu}_{\M{Z}_{\M{B}_K}}\! \bigr)
\end{align*}}To update each {\small $\M{Z}_{\M{B}_k}$}, we have to solve the above tri-diagonal system, for which we utilize Thomas' algorithm only using the scalar weights for efficiency. For the third mode factor (i.e., {\small$\M{D}_k$ for $k=1,...,K$}) , we use ADMM adjusted to include the ridge penalty in the update of the primal variable \citep[Supplementary Material]{parafac2-aoadmm}. The ADMM approach to the subproblem for the second mode (i.e., {\small$\M{B}_k$ for $k=1,...,K$}) is summarized in Algorithm \ref{admm-B}. The tPARAFAC2 optimization procedure is outlined in Algorithm \ref{tparafac2-aoadmm}. If required, further regularization penalties can be imposed on any of the factor matrices and if done so, ADMM can be utilized for each subproblem such as done by \cite{parafac2-aoadmm}. Although AO-ADMM lacks theoretical guarantees for convergence, we have observed convergence in the majority of the runs within experiments of this study, in line with prior empirical observations (Roald et al., 2022). We provide further discussion of its convergence behavior under our specific experimental settings in the supplementary material.

\subsection{Missing data} \label{subsec:tparafac2-missing}

Frequently, the input tensor {\small$\T{X}$} may be observed only partly. AO-ADMM algorithm for the regularized PARAFAC2 problem in \eqref{parafac2-aoadmm} has previously been only studied for fully observed data. Here, we consider two different ways of fitting the regularized PARAFAC2 model using AO-ADMM to incomplete data. The first is an Expectation Maximization (EM)-based approach, which imputes the missing entries, while the second one fits the model only to the observed entries, using row-wise (RW) updates. Once PARAFAC2 AO-ADMM is extended to incomplete datasets, we also use it to fit tPARAFAC2 to incomplete datasets.

\cite{parafac2_direct} initially proposed an EM-based ALS algorithm to fit PARAFAC2 to incomplete data. Missing entries are first imputed and then adjusted after a full model update based on the model estimates, following an EM-like approach \citep{DeLa77}. We incorporate this idea into the PARAFAC2 AO-ADMM framework of \cite{parafac2-aoadmm} and obtain one of the proposed approaches (that we refer to as AO-ADMM (EM) in the experiments) to handle missing data in a constrained PARAFAC2 model. An outline of this approach is shown in Algorithm \ref{alg-parafac2-em}. The algorithm alternates between imputing the missing entries with values from the model reconstruction (E-step) and updating the model parameters (M-step). The stopping conditions are identical to those mentioned for the PARAFAC2 AO-ADMM framework in the previous section. Algorithm \ref{alg-parafac2-em} can directly be extended to tPARAFAC2 models. Another related approach for handling missing data is to use ADMM with an auxiliary tensor variable that models the complete data. The factorization then approximates the auxiliary tensor, while the auxiliary tensor is fitted to the data tensor at the known entries.
This strategy has first been proposed in the AO-ADMM framework of \cite{aoadmm} for constrained CP models. It can be considered as a variation of the EM approach described above where the missing entries are imputed in each inner ADMM iteration instead of after one full outer AO iteration as in Algorithm \ref{alg-parafac2-em}. The approach has been extended to PARAFAC2 models with missing entries in REPAIR \citep{repair}, which addresses the additional problem of erroneous entries alongside missing data.

A different technique for fitting PARAFAC2 models to data with missing entries has been proposed in ATOM \citep{atom}. There, the model is fitted to the known entries only by utilizing row-wise updates for the factor matrices. Inspired by this work, we can reformulate the regularized PARAFAC2 in the presence of missing entries as:
{\small\begin{equation} \label{parafac2-aoadmm-rw}
\begin{split} 
     \min_{\M{A},\{\M{B}_k\}_{k=1}^K,\M{C}} \quad & \Bigg\{ \sum_{k=1}^{K} \fnorm{ \M{W}_k \Hada ( \M{X}_k \!-\! \M{A} \M{D}_k \M{B}_k^T)}^2 + g_{\M{A}}(\M{A})    \\ 
      & \quad + g_{\M{B}}(\{\M{B}_k\}_{k=1}^{K}) + g_{\M{D}}(\{\M{D}_k\}_{k=1}^{K}) \Bigg\} \\
    \text{subject to} \; \; \quad & \quad \{\M{B}_k\}_{k=1}^{K} \in  \mathscr{P} \quad  \forall \; k=1,\dots,K 
\end{split}
\end{equation}}where {\small$\T{W}$} is binary tensor of same size as {\small$\T{X}$} and
{\small
\begin{equation} \label{indicator}
    \T{W}(i,j,k) = \begin{cases}
    1 & \text{if } \T{X}(i,j,k) \text{ is observed,} \\
    0 & \text{otherwise.}
    \end{cases}
\end{equation}}Again, {\small$g_{\M{A}}$}, {\small $g_{\M{B}}$} and {\small $g_{\M{D}}$} denote regularization penalties imposed on the respective factors. The main difference with \eqref{parafac2-aoadmm} is that the model is now only fit to the observed entries.

\begin{algorithm}[t]
\caption{PARAFAC2 AO-ADMM EM}\label{alg-parafac2-em}
\textbf{Input:} Data input {\small$\T{X}$}, Indicator tensor {\small$\T{W}$} \algorithmiccomment{Equation \eqref{indicator}}\\
\textbf{Output:} PARAFAC2 factors {\small$\M{A},\{\M{B}_k\}_{k=1}^{K},\{\M{D}_k\}_{k=1}^{K}$}.
\begin{algorithmic}[1]
\State \textit{Initialize} {\small$\M{A}, \M{Z}_{\M{A}}, \M{\mu}_{\M{A}} $}
\State \textit{Initialize} {\small$\{\M{B}_k,\!\M{Y}_{\M{B}_k},\M{Z}_{\M{B}_k},\M{\mu}_{\M{Z}_{\M{B}_k}},\M{\mu}_{\M{\Delta}_{\M{B}_k}}\}_{k=1}^{K}$}
\State \textit{Initialize} {\small$\{\M{D}_k,\M{Z}_{\M{D}_k},\M{\mu}_{\M{D}_k}\}_{k=1}^{K} $}
\State \textit{Initialize}  {\small$\T{X}(\T{W}=0)$} \algorithmiccomment{Missing entries of  {\small$\T{X}$}}
\State \textbf{while} \textit{outer stopping conditions are not met}
\State \hspace{0.25cm} {\small$\{\M{B}_k, \M{Y}_{\M{B}_k}, \M{Z}_{\M{B}_k}, \M{\mu}_{\M{Z}_{\M{B}_k}}, \M{\mu}_{\M{\Delta}_{\M{B}_k}}\}_{k=1}^{K} \gets \text{\cite[Algorithm 4]{parafac2-aoadmm}}$}
\State \hspace{0.25cm} {\small$\{{{\M{D}_k},\M{Z}_{\M{D}_k}},\M{\mu}_{\M{D}_k}\}_{k=1}^K \gets \text{\cite[Algorithm 7]{parafac2-aoadmm}}$}
\State \hspace{0.25cm} {\small$\{\M{A}, \M{Z}_{\M{A}}, \M{\mu}_{\M{A}}\} \gets \text{\cite[Algorithm 6]{parafac2-aoadmm}}$}
\State \hspace{0.25cm} {\small $\hat{\T{X}}$} $\gets$ {\small $\textit{reconstruct}(\M{A}, \{\M{B}_k\}_{k=1}^{K}, \{\M{D}_k\}_{k=1}^{K})$}
\State \hspace{0.25cm} {\small $\T{X}(\T{W}=0)$} $\gets$ {\small $\hat{\T{X}}(\T{W}=0)$}
\State \textbf{end} 
\end{algorithmic}
\end{algorithm}

Since we cannot solve \eqref{parafac2-aoadmm-rw} for any full factor matrix, we resolve to row-wise updates. After forming the augmented Lagrangian {\small $\mathcal{L}$} of \eqref{parafac2-aoadmm-rw}, all update rules in the PARAFAC2 AO-ADMM framework \citep{parafac2-aoadmm} that do not involve the fidelity term remain identical and valid. Hence, we only need to adjust our approach to solving the subproblems for each of the factor matrices. Isolating the relevant terms of {\small $\mathcal{L}$} for each row of each factor matrices yields:

{\small
\begin{equation}
\begin{split}  \label{a-rw-rw}
    \min_{\M{A}(i,:)} \  \Bigg\{ \sum_{k=1}^{K} & \fnorm{\M{W}_k(i,:)   \Hada \M{X}_k(i,:) - \M{W}_k(i,:) \Hada (\M{A}(i,:) \M{D}_k \M{B}_k\Tra)}^2  \\
    & + \frac{\rho_A}{2} \fnorm{\M{A}(i,:) - \M{Z}_A(i,:) + \M{\mu}_{\M{A}}(i,:)}^2 \Bigg\} \ \  \forall i=1,...,I
\end{split}
\end{equation}}
{\small
\begin{equation}
\begin{split}  \label{b-rw-rw}
    \min_{\M{B}_k(j,:)} \  \Bigg\{ & \lVert diag(\M{W}_k(:,j))  \M{X}_k(:,j) - diag(\M{W}_k(:,j)) (\M{A} \M{D}_k \M{B}_k(j,:)\Tra) \rVert_{F}^2 \\
    & + \frac{\rho_{\M{B}_k}}{2} \fnorm{\M{B}_k(j,:) - \M{Y}_{\M{B}_k}(j,:) + \M{\mu}_{\M{Z}_{\M{\Delta}_{\M{B}_k}}}(j,:)}^2 \\
    & + \frac{\rho_{\M{B}_k}}{2} \fnorm{\M{B}_k(j,:) - \M{Z}_{\M{B}_k}(j,:) + \M{\mu}_{\M{Z}_{\M{B}_k}}(j,:)}^2 \Bigg\} \\
    & \quad \forall j=1,...,J \quad \forall k=1,...,K \\
\end{split}
\end{equation}}
{\small
\begin{equation}
\begin{split}  \label{c-rw-rw}
    \min_{\M{D}_k} \  \Bigg\{ \sum_{\substack{i, j \\ \M{W}_k(i,j) = 1}}  & (\M{X}_k(i,j) - (\M{B}_k(:,j) \Khat \M{A}(i,:))diag(\M{D}_k)\Tra )^{2} \\
    & + \frac{\rho_{\M{D}_k}}{2} \fnorm{\M{D}_k - \M{Z}_{\M{D}_k} + \M{\mu}_{\M{D}_k}}^2 \Bigg\} \\
    & \quad \forall k=1,...,K
\end{split} 
\end{equation}}To find the minimizers of Equations \eqref{a-rw-rw}, \eqref{b-rw-rw}, \eqref{c-rw-rw}, we set the respective partial derivatives equal to zero and obtain the update rules shown in \eqref{rw-update-a}, \eqref{rw-update-b} and \eqref{rw-update-c} (for brevity, we use {\small$\M{C}(k,:)$} instead of {\small$diag(\M{D}_k)$} in \eqref{rw-update-c}). Replacing the factor update rules in the PARAFAC2 AO-ADMM framework with \eqref{rw-update-a}, \eqref{rw-update-b} and \eqref{rw-update-c} (for each factor row) allows us to solve \eqref{parafac2-aoadmm-rw} regardless of the type of imposed regularization, including the temporal regularization of the proposed tPARAFAC2. We refer to this approach as  AO-ADMM (RW) in the experiments. We also highlight the fact that such updates are easily parallelizable.

{\small
    \begin{equation} \label{rw-update-a}
    \begin{split}
        \M{A}^{*}(i,:) \gets & \left( \sum_{k=1}^{K} \M{X}_k(i,:)\M{B}_k\M{D}_k 
        + \frac{\rho_{A_i}}{2} \left( \M{Z}_{A}(i,:)-\M{\mu}_{A}(i,:) \right) \right) \\
        & \times \left( \sum_{k=1}^{K} \M{D}_k \M{B}_k\Tra \, \text{diag}(\M{W}_k(i,:)) \M{B}_k \M{D}_k 
        +  \frac{\rho_{\M{A}_i}}{2} \M{I}_{R} \right)^{-1}
    \end{split}
    \end{equation}

    \begin{equation} \label{rw-update-b}
    \begin{split}
        \M{B}_k^{*}(j,:) \gets & \left( \M{X}_k(:,j)\Tra \M{A} \M{D}_k \right. \\
        & + \frac{\rho_{B_{k,j}}}{2} \left( \M{Z}_{\M{B}_k}(j,:) - \M{\mu}_{\M{Z}_{\M{B}_k}}(j,:) \right. \\
        & \left. \left. \quad + \M{Y}_{\M{B}_k}(j,:) - \M{\mu}_{\M{\Delta}_{B_k}}(j,:) \right) \right) \\
        & \times \left( \M{D}_k \M{A}\Tra \, \text{diag}(\M{W}_k(:,j)) \M{A} \M{D}_k 
        + \rho_{B_{k,j}} \M{I} \right)^{-1}
    \end{split}
    \end{equation}

    \begin{equation} \label{rw-update-c}
    \begin{split}
        \M{C}^{*}(k,:) \gets & \left( \sum_{\substack{i, j \\ \M{W}_k(i,j) = 1}}
        \left( \M{A}(i,:)\Tra \M{A}(i,:) \Hada \M{B}_k(j,:)\Tra \M{B}(j,:) \right) \right. \\
         \left. \quad + \frac{\rho_{{D}_k}}{2} \M{I}_{R} \right)^{-1} & \times \left( \text{diag}(\M{A}\Tra \M{X}_k \M{B}_k) 
        + \frac{\rho_{{D}_k}}{2} \left( \M{Z}_{\M{D}_k} - \M{\mu}_{\M{D}_k} \right) \right)
    \end{split}
    \end{equation}

}Although the literature contains comparable methods \citep{repair,atom}, they differ from the proposed methods. REPAIR \citep{repair} employs an EM-related approach for missing data and uses ADMM for regularizing some factor matrices. However, since it uses the reparametrization of the PARAFAC2 constraint as in \eqref{parafac2-direct}, no regularization can be imposed on factor matrices {\small$\M{b}_k$}. ATOM \citep{atom} utilizes row-wise factor updates, but the approach also adapts the flexible PARAFAC2 constraint \citep{parafac2-flexiblecoupling} which is not flexible enough to adapt to different kinds of regularization penalties and involves additional hyperparameters.

\section{Numerical Experiments} \label{sec:experiments}

\subsection{Experimental setup}

\begin{figure*}[t]
    \centering
    \subfloat[]{\includegraphics[width=0.28\textwidth]{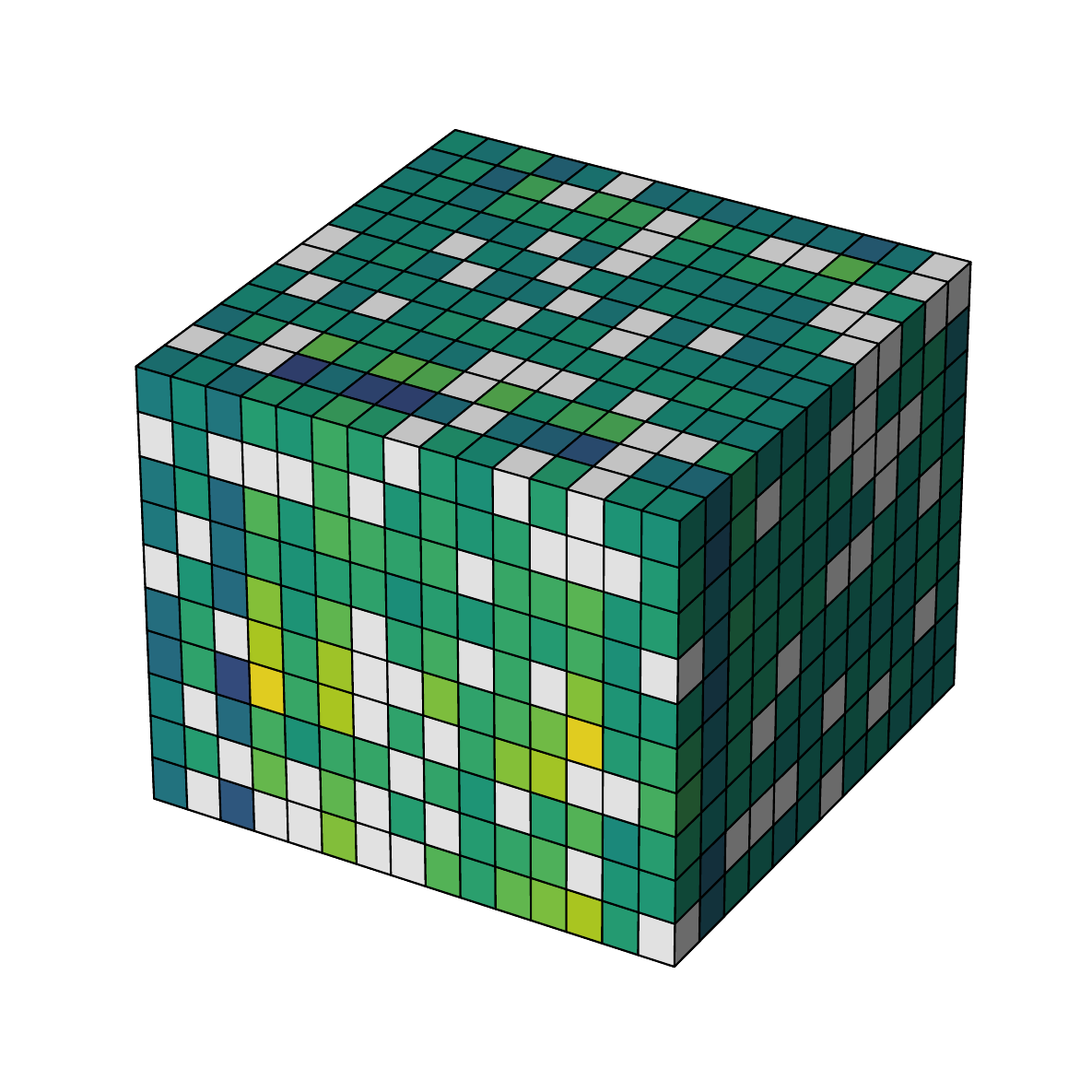}\label{fig:sub1}}
    \hfill
    \subfloat[]{\includegraphics[width=0.28\textwidth]{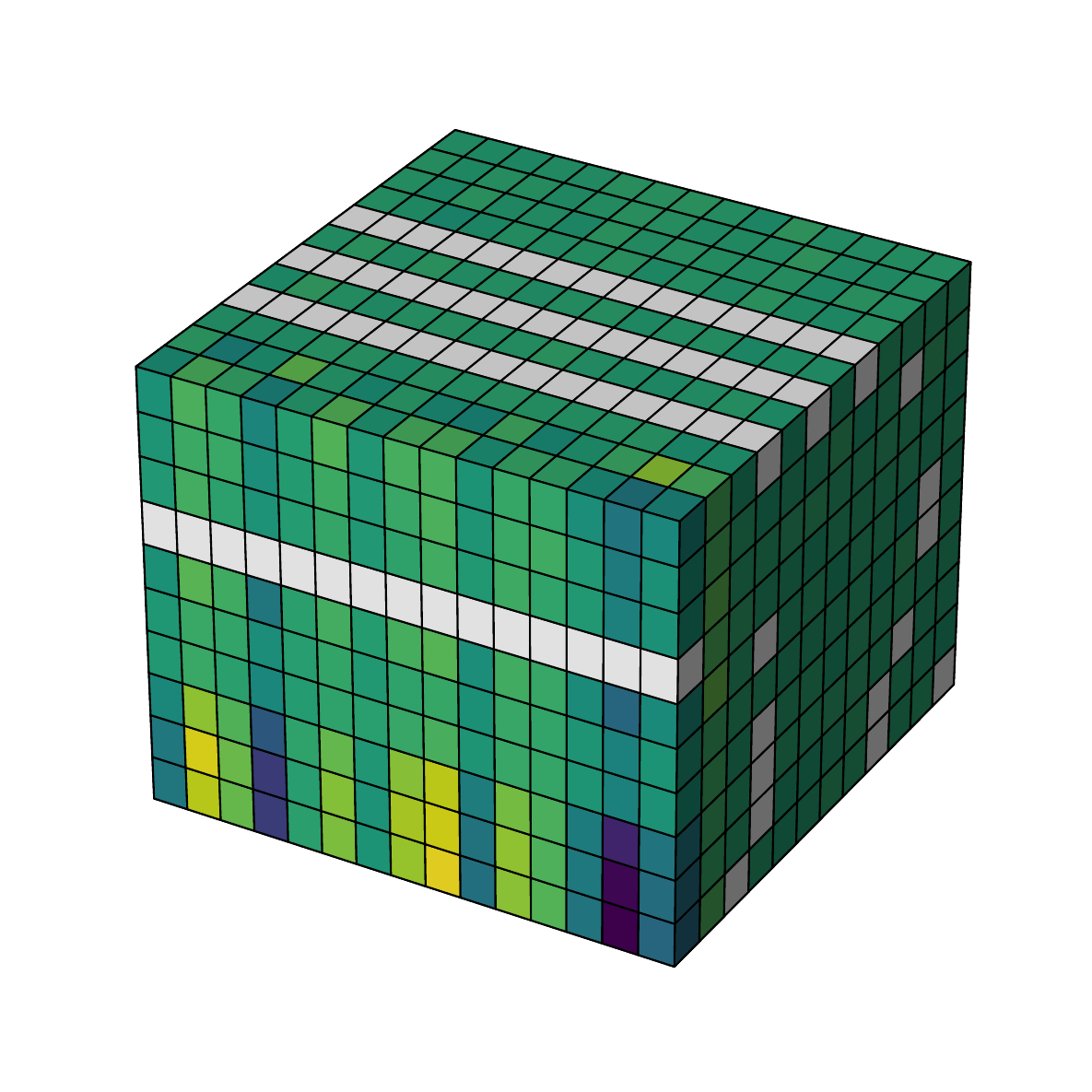}\label{fig:sub2}}
    \hfill
    \subfloat[]{\includegraphics[width=0.28\textwidth]{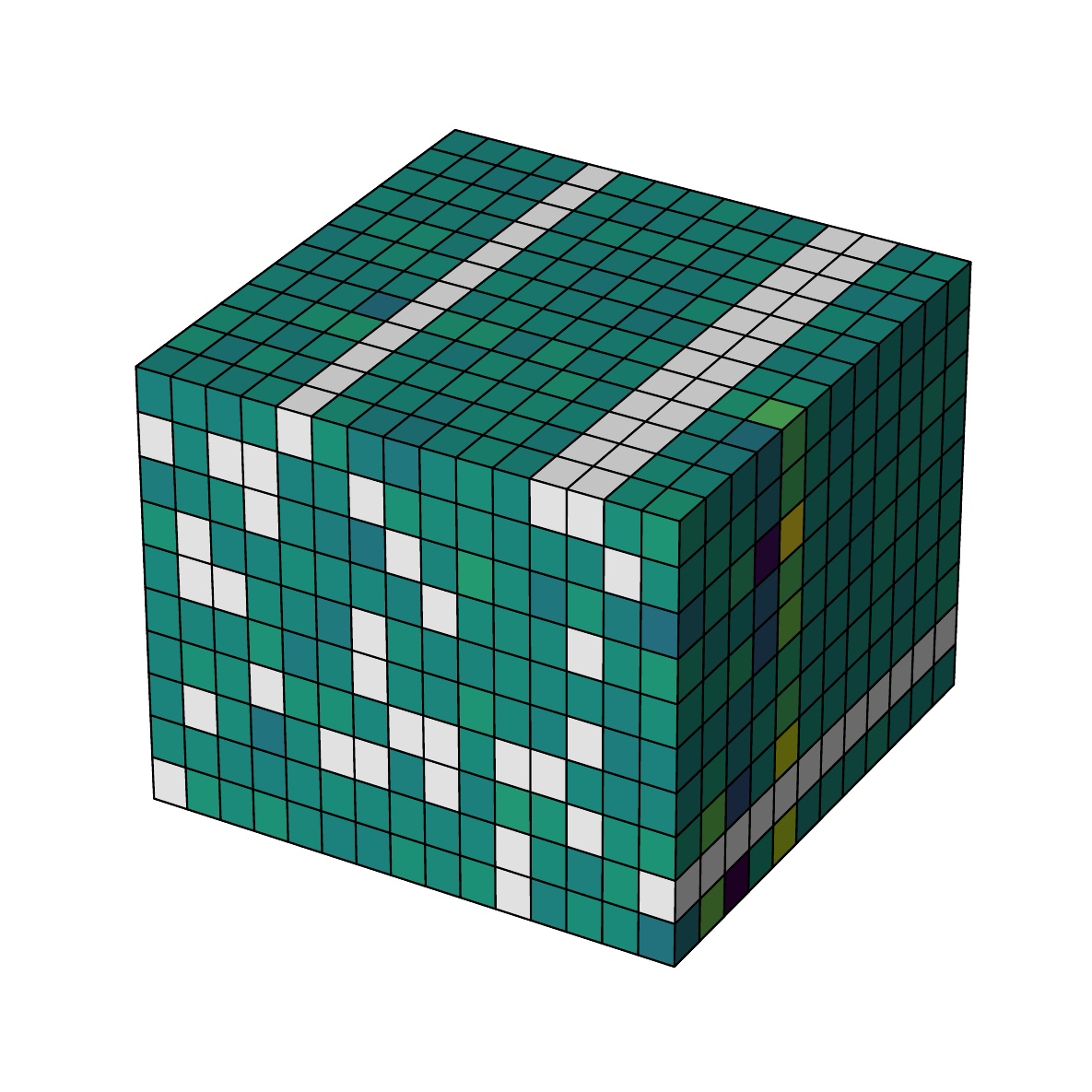}\label{fig:sub3}}
    \caption{Illustrations of missing data settings considered on a third-order tensor. Cubes of white color denote the respective entry is missing. In setting (a), missing entries have no particular structure (section \ref{sec:random}), in (b) and (c) mode-2 and mode-3 fibers are missing (section \ref{sec:structured}).}
    \label{fig:missingness_settings}
\end{figure*}

In this section, we evaluate tPARAFAC2 and the two different ways of handling missing data (i.e., AO-ADMM (EM) and AO-ADMM (RW)) using simulations and real-world datasets. All synthetic datasets contain slowly changing patterns as in Figure \ref{fig:sample_incremental}. We first assess the performance of tPARAFAC2 on fully observed data with different noise levels in terms of capturing the evolving patterns accurately. As baselines, we consider PARAFAC2, PARAFAC2 with ridge penalties in all modes, and the less structurally constrained model t(emporal)CMF formulated as follows based on the CMF formulation in \cite{chimera}:
{\small
\begin{equation} \label{tcmf-abstract}
\begin{split} 
     \min_{\M{A},\{\M{B}_k\}_{k=1}^K} \quad & \Bigg\{ \sum_{k=1}^{K} \fnorm{\M{X}_k \!-\! \M{A} \M{B}_k^T}^2 + \lambda \fnorm{\M{A}}^2 \\
     & + \lambda_\M{B} \sum_{k=2}^K\fnorm{ \! \M{B}_k \!-\! \M{B}_{k-1} \! }^{2} \Bigg\} \ .
\end{split}
\end{equation}}We then study the performance of PARAFAC2 and tPARAFAC2 in terms of revealing the evolving patterns from incomplete datasets (with random and structured missing data) by comparing the two different approaches of handling missing data. The settings considered for missing data are illustrated in Figure \ref{fig:missingness_settings}. Finally, we demonstrate on a chemometrics dataset that the extension of AO-ADMM algorithm for regularized PARAFAC2 to incomplete data can recover the underlying patterns accurately in the presence of missing data. Using a metabolomics dataset, we also demonstrate the performance of PARAFAC2, PARAFAC2 and tCMF in terms of revealing evolving patterns and the effectiveness of temporal regularization in the presence of high amounts of missing data.

In all cases, we use the Factor Match Score (FMS) as a measure of accuracy, which is defined as follows:
{\small\begin{equation}
    \text{FMS} = \frac{1}{R} \sum_{i=1}^R\frac{|\MhatC{A}{i}\Tra\MC{A}{i}|}{\lVert \MhatC{A}{i} \rVert \lVert \MC{A}{i} \rVert} \frac{|\MhatC{B}{i}\Tra\MC{B}{i}|}{\lVert \MhatC{B}{i} \rVert \lVert \MC{B}{i} \rVert}  \frac{|\MhatC{C}{i}\Tra\MC{C}{i}|}{\lVert \MhatC{C}{i} \rVert \lVert \MC{C}{i} \rVert},
\end{equation}}where {\small$\MC{A}{i},\MC{B}{i}$} and {\small $\MC{C}{i}$} denote the {\small$i$}-th column of ground truth factors, and {\small $\MhatC{A}{i},\MhatC{B}{i}$}, {\small $\MhatC{C}{i}$} are their model estimates. {\small $\MhatC{B}{i}$} and {\small$\MC{B}{i}$} refer to vectors produced by stacking the {\small$i$}-th column of all {\small$\{\M{B}_k\}_{k=1}^{K}$} matrices. FMS is computed after aligning the order of the components for both decompositions. When comparing tCMF factors, we first column normalize each column of the evolving factors separately and ``form" the {\small $\M{C}$} matrix by absorbing these entries, i.e. {\small $\M{C}(k,r)=\fnorm{\M{B}_k(:,r)}$}. In addition, since evolving patterns have zeros (corresponding to inactive parts), to better quantify the error on the evolving factors, we  also use root mean squared error for the evolving mode, i.e., $\text{RMSE}_B$, defined as:
{\small\begin{equation} \label{rmse}
    \text{RMSE}_B = \sqrt{\sum_{k=1}^K \frac{\sum_{j,r}( {\small \M{B}_k}(j,r) - \hat{\M{B}_k}(j,r))^2}{K J R}}
\end{equation}}where {\small $\M{B}_k$} is the ground truth factor, {\small $\hat{\M{B}_k}$} its model estimate and {\small $K J R$} is the total number of elements considered, assuming the input of {\small $\T{X}\in\Real^{I \times J \times K}$. {\small $\text{RMSE}_B$} is in all cases computed after appropriately permuting the components and matching the signs between each column of {\small $\M{B}_k$} and {\small $\hat{\M{B}_k}$}. }

Implementation of all methods is based on TensorLy \citep{tensorly} and MatCoupLy \citep{matcouply}, and our code can be found here\footnote{\href{https://github.com/cchatzis/tPARAFAC2-latest}{https://github.com/cchatzis/tPARAFAC2-latest}}. For all experiments and methods, the maximum number of iterations is set to {\small $10000$}, the (relative) tolerance of outer AO-ADMM loops to {\small$10^{-8}$}, inner loop tolerances to {\small$10^{-5}$}. More information on stopping conditions is given in the supplementary material. We only take into account runs with `feasible' solutions with a tolerance of {\small$10^{-5}$} (i.e. constraints are allowed to be violated at most this much). Experiments were performed on Ubuntu 22.04 on AMD EPYC 7302P 16-core processors. Time measurements were taken using Python's \textit{time} module.

\subsection{Simulated data analyis}

\begin{figure*}[t]
    \begin{minipage}[b]{1.0\linewidth}
      \centering
      \centerline{\includegraphics[width=1.0\textwidth]{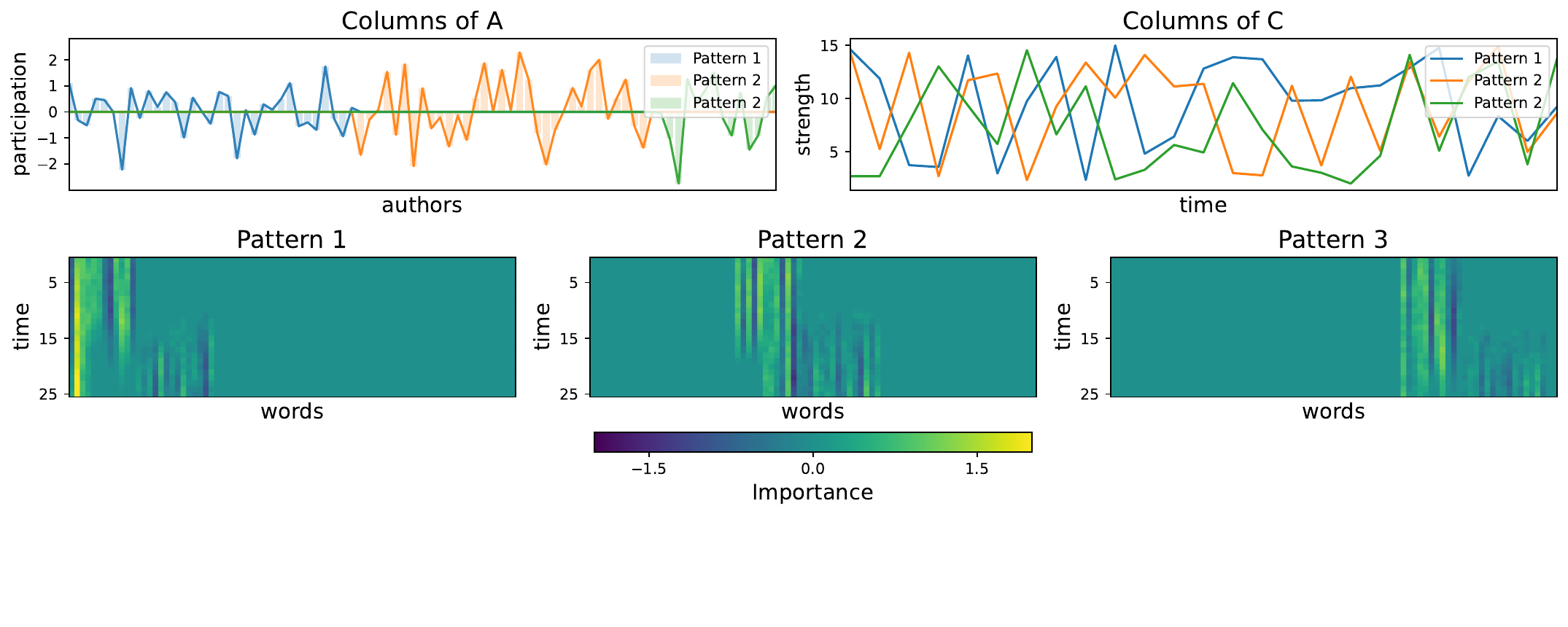}}
      \vspace{-1.5cm}
      \caption{One of the simulated datasets. The columns of {\small$\M{A}$} capture the authors partaking in each concept, while the columns of {\small$\M{C}$} reflect the pattern strength at each time point. Each heatmap shows how the column of {\small$\{\M{B}_k\}_{k=1}^{25}$}  that corresponds to a specific pattern changes in time ($\forall k=1,...,25)$. }
    \label{fig:sample_incremental}
    \end{minipage}
\end{figure*}

We first assess the performance of tPARAFAC2 on synthetic data. One potential use case would be to track evolving concepts across an $authors \times words \times time$ tensor. To generate such data, we generate ground truth factors {\small$\M{A},\{\M{B}_k\}_{k=1}^{K}$} and {\small$\{\M{D}_k\}_{k=1}^{K}$} that reflect the evolution of three concepts across time, and then form each frontal slice of the dataset as {\small$\M{X}_k = \M{A} \M{D}_k \M{B}_k\Tra, \  \forall k=1,..., K$ }. Figure \ref{fig:sample_incremental} shows factors of such a dataset. Each concept consists of three parts: 
\begin{enumerate}
    \item[\itshape Authors:] A list of relevant authors participating in the concept. We simulate that by selecting a set of relevant author-indices for the respective column of factor {\small$\M{A}$} and drawing `participation' values from {\small$\mathcal{N}(0,1)$}, while the rest is set to $0$.
    \item[\itshape Words:] The words that compose each concept across time. This is captured by the non-zero indices of the respective column across all {\small$\{\M{B}_k\}_{k=1}^{K}$} factors. For each concept, we randomly choose an initial and final word set, in which the concept has non-zero values at the first and last time step, respectively. The `importance' values for the initial set are drawn from {\small$\mathcal{N}(0,1)$}. To model smooth evolution, we incrementally add a value from {\small$\mathcal{N}(0,0.1)$} over time. After a randomly chosen time point, the concept begins to shift towards the final word set: with a probability of $0.3$, a word that is in the initial set but not in the final will start reducing to zero, or a new word from the final set will be initialized with a value from {\small$\mathcal{N}(0,0.1)$} or both events occur. {\small$30\%$} of the words of each concept will remain active at all times, as they are chosen to be in both the initial and final set.
    \item[\itshape Popularity:] The pattern's strength (popularity) across time, reflected in {\small$\M{C}$}. We simulate that by drawing values from {\small$\mathcal{U}(1,15)$}. We make sure the congruence coefficient, i.e., the cosine similarity between columns, is no more than {\small 0.8} in each generated dataset \citep{parafac2_direct}.
\end{enumerate}

\subsubsection{Different noise levels} Here, we assess the performance of tPARAFAC2 in terms of accuracy on data with slowly changing patterns in the presence of different amounts of noise. We create 20 datasets of size {\small$100\times80\times25$} that contain three concepts. After forming the data tensor {\small$\T{X}$}, we add noise as follows:
{\small\begin{equation}
    \T{X}_{noisy} = \T{X} + \eta \lVert \T{X}\rVert \frac{\Theta}{\lVert \Theta \rVert}
\end{equation}}where {\small$\Theta\sim \mathcal{N}(0,1)$} and {\small$\eta\in\{0.5,1.0,1.5,2.0\}$} controlling the noise level. We compare PARAFAC2, PARAFAC2 with ridge regularization on all modes, tCMF and tPARAFAC2 in terms of recovering the ground truth factors. Three components are used for all methods (i.e., {\small$R=3$}), since we know a priori the number of underlying patterns. Non-negativity is imposed on {\small $\M{C}$} in PARAFAC2-based models to alleviate the sign ambiguity, and AO-ADMM is used to fit the models. For each dataset, thirty random initializations are generated that are used by all methods. After discarding degenerate runs \citep{degen}, we choose the best run for each method for each dataset according to the lowest loss function value. Runs that reached the maximum number of iterations or yielded unfeasible solutions are not considered. 

\begin{figure}[t]
    \begin{minipage}{1.0\linewidth}
      \centering
      \centerline{\includegraphics[width=\textwidth]{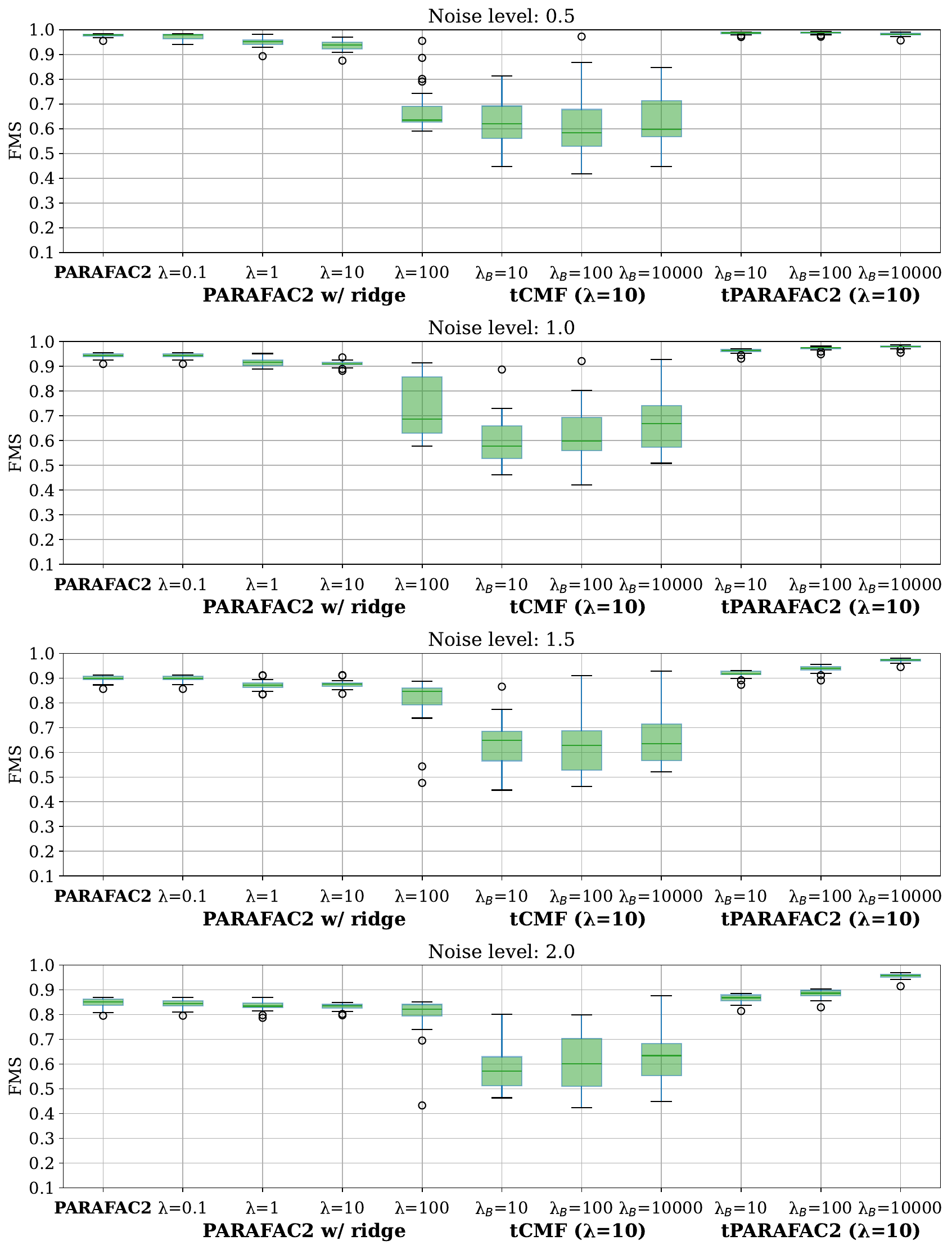}}
      \caption{FMS for all methods at different noise levels when compared with the ground truth factors. Each boxplot contains 20 points, one for the best-performing run of the method in each dataset. {\small $\lambda$} denotes the strength of the ridge penalty imposed on all factors in the PARAFAC2 model, specifically to factors {\small $\M{A}$} and {\small $\{{\M{D}}\}_{k=1}^{25}$} in the tPARAFAC2 model, and only to factor {\small $\M{A}$} in the tCMF model. {\small $\lambda_B$} represents the strength of the temporal smoothness penalty.}
    \label{fig:incrementals}
    \end{minipage}
\end{figure}

\begin{figure}[t]
    \begin{minipage}{1.0\linewidth}
      \centering
      \centerline{\includegraphics[width=\textwidth]{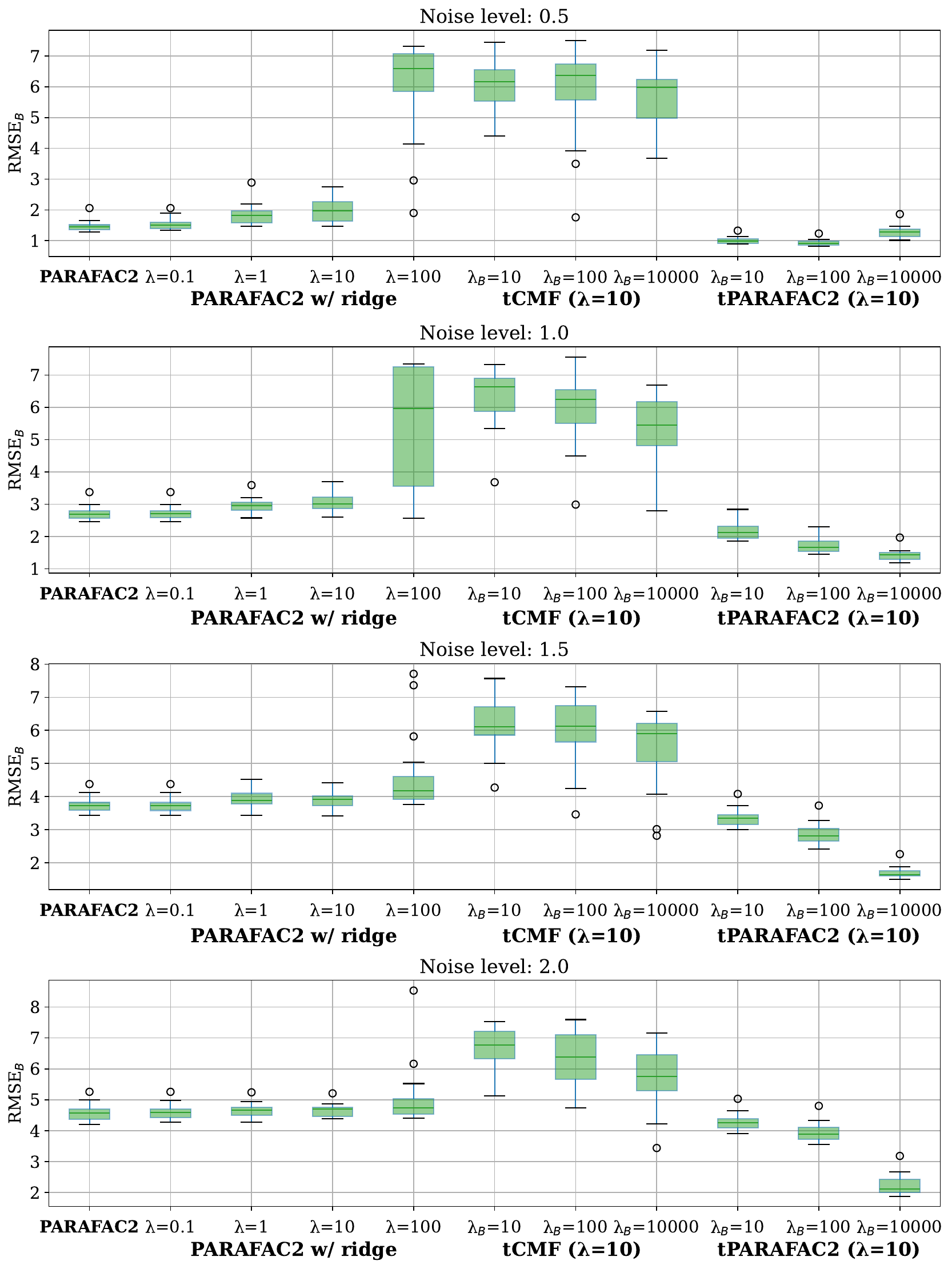}}
      \caption{$\text{RMSE}_B$ of all methods at different noise levels. Each boxplot contains 20 points, one for the best-performing run of the method in each dataset.}
    \label{fig:rmse}
    \end{minipage}
\end{figure}

\begin{figure}[t]
    \centering
    \subfloat[]{%
        \includegraphics[width=\columnwidth]{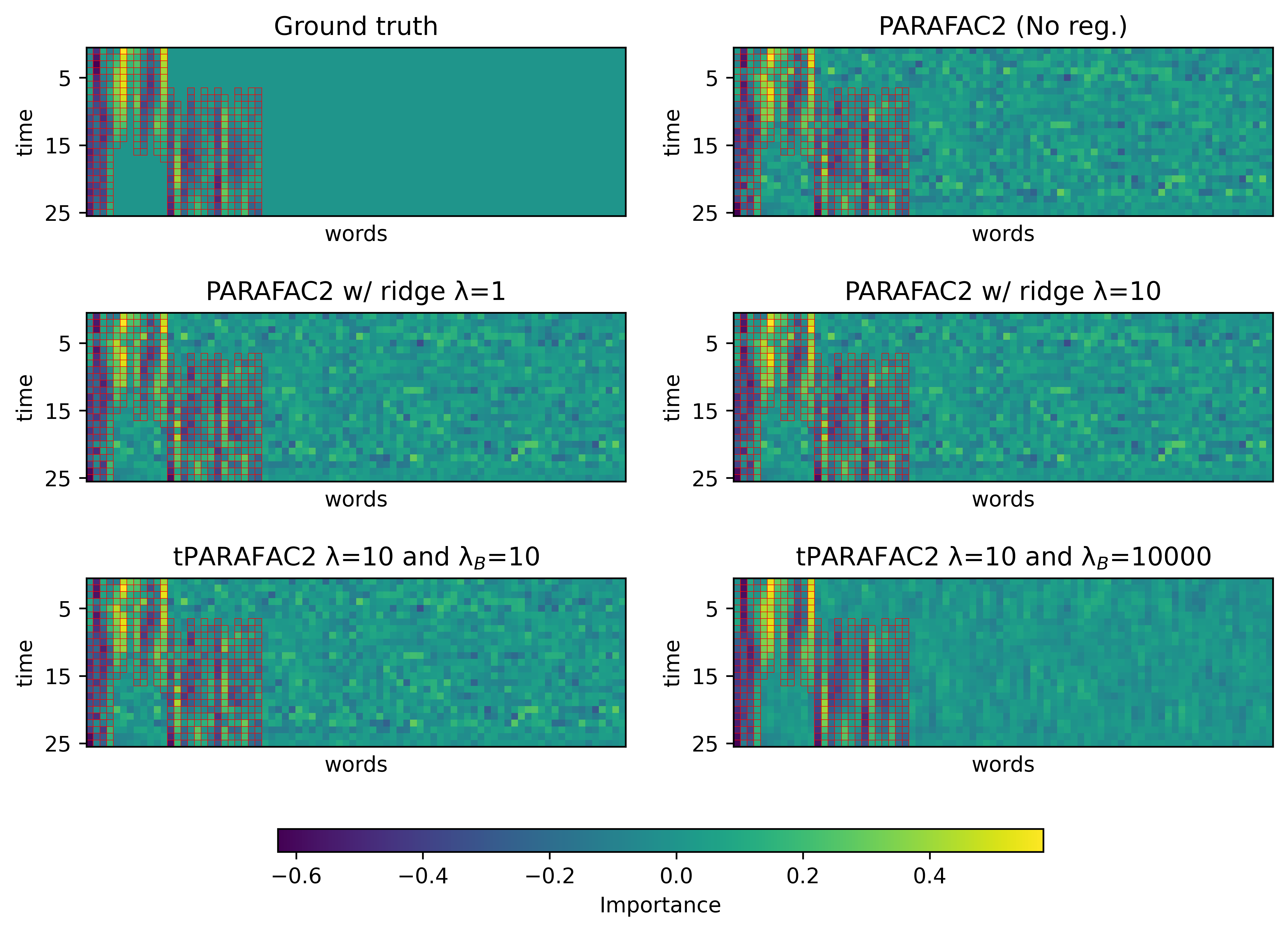}
        \
    }

    \subfloat[]{%
        \begin{tabular}{|c|c|c|c|}
        \hline
        \multirow{2}{*}{\shortstack{{\small\textbf{Method}}}} & \multirow{2}{*}{\shortstack{{\small\textbf{Cosine sim.}} \\ {\small\textbf{of `active'}}}} & \multirow{2}{*}{\shortstack{{\small\textbf{Norm of}} \\ {\small\textbf{`inactive'}}}} & \multirow{2}{*}{\shortstack{{\small\textbf{RMSE}$_B$}}} \\
        & & &\\
        \hline
        \hline
        {\small PARAFAC2} & 0.90 & 2.81  & 0.081 \\
        \hline
        \multirow{2}{*}{{\small \shortstack{PARAFAC2 w/\\ ridge $\lambda = 1$}}} & \multirow{2}{*}{0.90} & \multirow{2}{*}{2.79} & \multirow{2}{*}{0.080} \\ 
        & & &\\
        \hline
        \multirow{2}{*}{{\small \shortstack{PARAFAC2 w/\\ ridge $\lambda = 10$}}} & \multirow{2}{*}{0.90} & \multirow{2}{*}{2.79} & \multirow{2}{*}{0.080} \\
        & & &\\
        \hline
        \multirow{2}{*}{{\small \shortstack{tPARAFAC2\\ $\lambda = 10$, $\lambda_{B}=10$}}} & \multirow{2}{*}{0.91} & \multirow{2}{*}{2.73} & \multirow{2}{*}{0.078} \\
        & & &\\
        \hline
        \multirow{2}{*}{{\small \shortstack{tPARAFAC2\\ $\lambda = 10$, $\lambda_{B}=10000$}}} & \multirow{2}{*}{0.98} & \multirow{2}{*}{1.66} & \multirow{2}{*}{0.046} \\
        & & &\\
        \hline
        \end{tabular}
    }
    \caption{(a) depicts a slowly changing pattern, with the `active' words marked in red. The second column of (b) shows the cosine similarity 
    between vectors of `active' words and the ground truth. The third column measures the norm of all non-active entries. The last column measures the RMSE of all entries with the ground truth (only the shown component is considered). These measurements are taken after normalizing the factors, with noise level {\small$\eta=2.0$}.}
    \label{fig:improvement}
\end{figure}

Figures \ref{fig:incrementals} and \ref{fig:rmse} show that while PARAFAC2 and tPARAFAC2 have comparable performance at low amounts of noise, tPARAFAC2 is able to recover the ground truth factors more accurately as the noise level increases. We omit higher hyperparameters for ridge regularization due to a consistent decline in accuracy. For tPARAFAC2, we noticed that the performance is sensitive to the hyperparameter chosen. In the supplementary material, we present a more complete sensitivity analysis where it can be seen that using lower values results in under-regularization, while using too high values oversmooth the factors. We also observed that higher noise levels required larger values of this hyperparameter to obtain improvements in accuracy. tCMF yields non-unique output, which explains the poor results. The main source of improvement for tPARAFAC2 is the increased accuracy of recovering {\small$\M{B}_k$} factor matrices (also confirmed by Figure \ref{fig:rmse}). An indicative example of why this is the case is shown in Figure \ref{fig:improvement}. To illustrate this, we first concatenate all non-zero elements of the ground truth concept into a vector and then compute the cosine similarity with the vectors containing the respective entries from the reconstructions, as shown in Figure \ref{fig:improvement}. Additionally, we estimate the noise remaining in the factor by measuring the norm of the rest of the entries (i.e., non-active). Notice that the cosine similarity cannot be used here since the ground truth is a zero vector. A high norm indicates a noisy factor. We can see that tPARAFAC2 is more accurate because the temporal smoothness (a) helps recover the ground truth structure more accurately and (b) makes the method more robust to noise. $\text{RMSE}_B$ shows the overall error reduction achieved by tPARAFAC2 when the right temporal smoothness parameter is used.

\subsubsection{Randomly missing entries} \label{sec:random}In this experiment, we assess the performance of tPARAFAC2 in terms of finding underlying slowly changing patterns in the presence of missing data. We construct 20 datasets with size {\small$100\times80\times25$} and add noise with {\small$\eta={0.75}$}. For each of those datasets, we create {\small$10$} binary tensor masks that indicate whether an entry is observed or not, with {\small$25\%$}, {\small$50\%$} and {\small$75\%$} entries set as missing. We compare the performace of PARAFAC2, PARAFAC2 with ridge regularization on all modes with strength {\small$\lambda\in\{0.1,1,10,100\}$} and tPARAFAC2 with ridge hyperparameters being {\small$\lambda=10$} and {\small$\lambda_B\in\{10,100,10000\}$}. Similar to the fully observed setting, we choose these values after performing a sensitivity analysis to avoid excessive or insufficient regularization. We consider AO-ADMM (EM) and AO-ADMM (RW) for fitting these models. As another baseline, we also use the standard PARAFAC2-ALS (EM) approach for handling missing data \citep{parafac2_direct}. For each mask, {\small$30$} random initializations (the same 30 across all methods) are used. For EM-based approaches, we use the mean of the known entries of each frontal slice as the initial estimate of missing entries.

\begin{figure}[t]
\begin{minipage}[b]{\linewidth}
\centering      \centerline{\includegraphics[width=\textwidth]{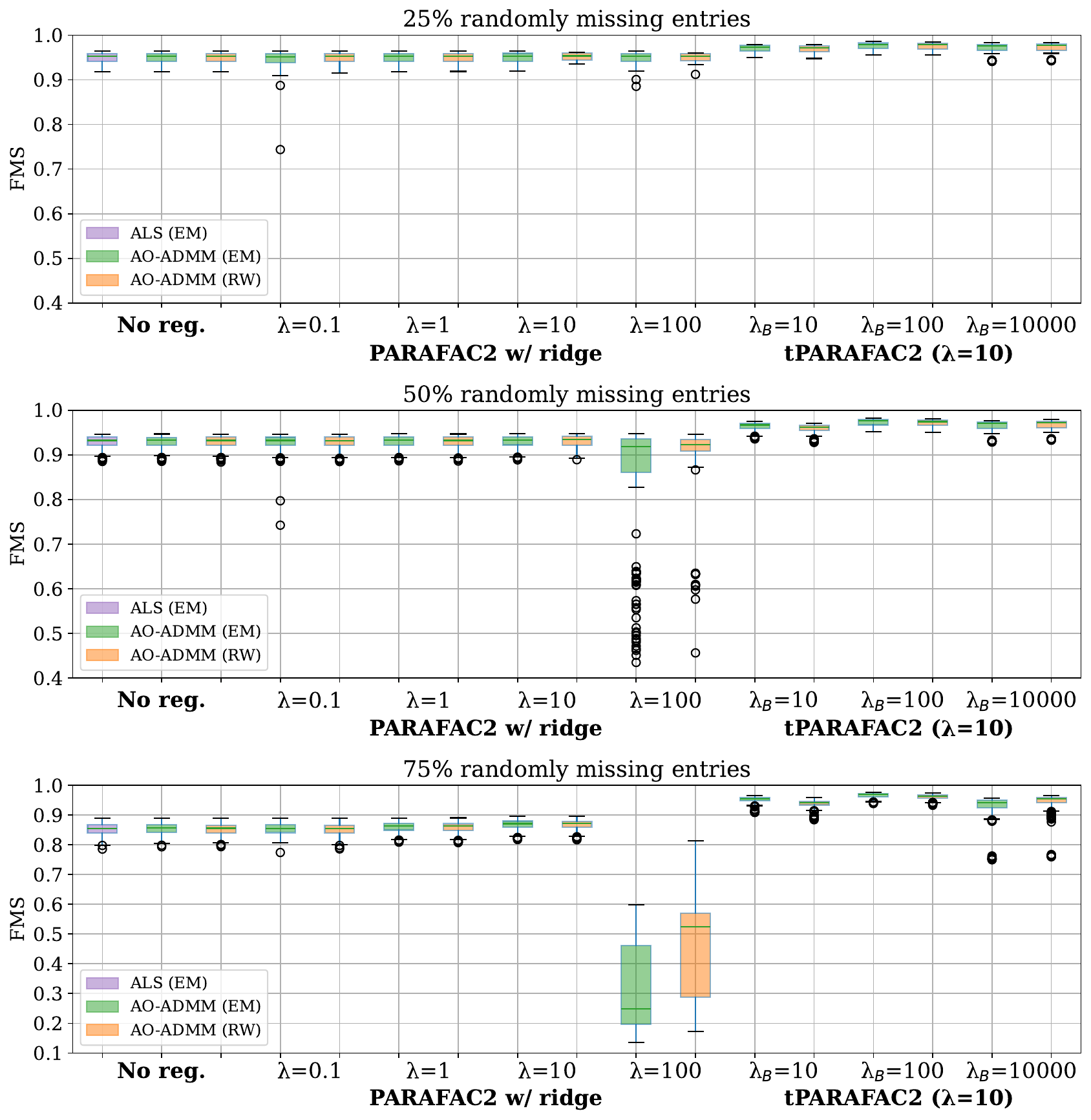}}
\caption{Accuracy of methods in terms of FMS with the ground truth, when the input contains randomly missing entries. Each boxplot contains {\small$200$} points, one for the best-performing run of the method at each mask of each dataset.}
\label{fig:random_missing}
\end{minipage}
\end{figure}

\begin{figure}[t]
    \centering
    \subfloat[]{%
    \includegraphics[width=\columnwidth]{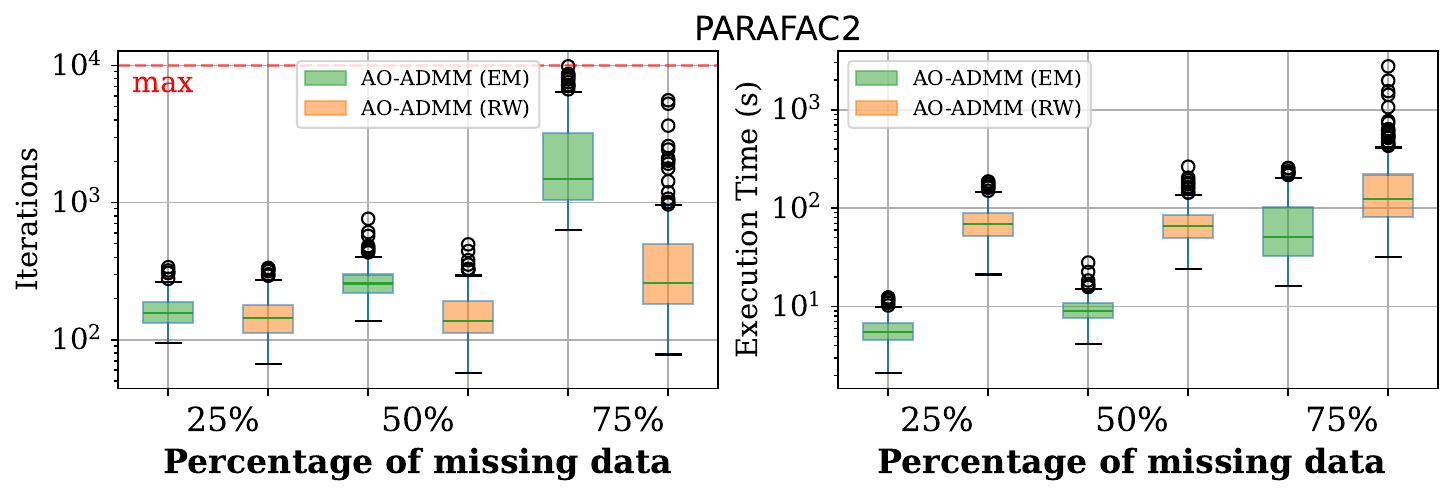}
    \label{fig:comparison1}
    }

    \subfloat[]{%
        \includegraphics[width=\columnwidth]{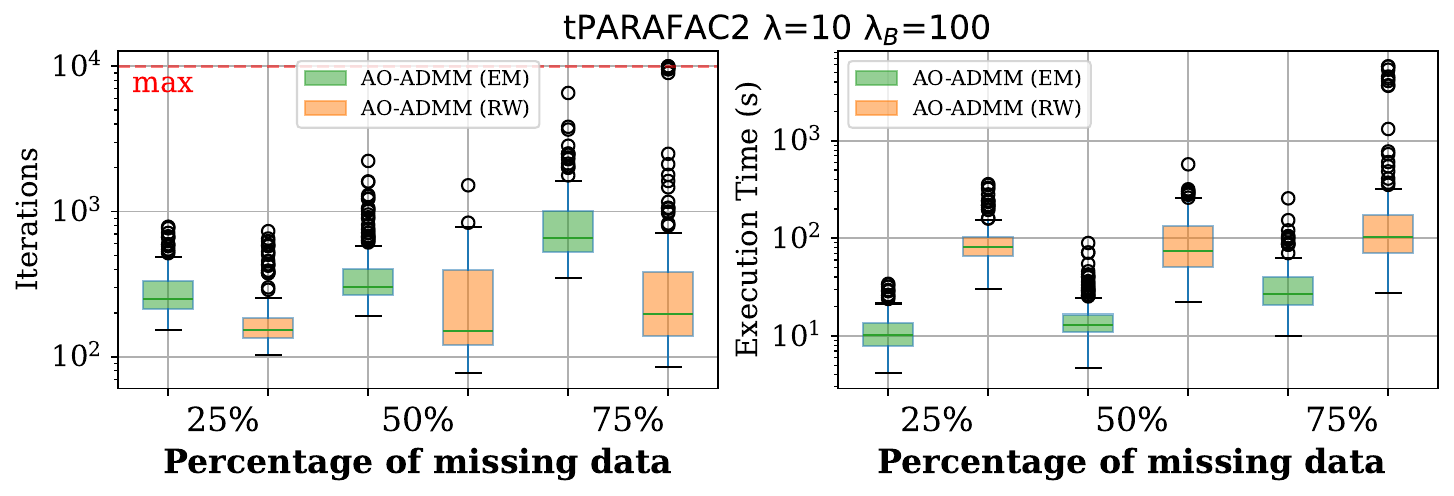}
        \label{fig:comparison2}
    }
      \caption{Comparison of EM and RW methods in terms of computation time. The measurements are taken on the runs of the \textit{randomly missing entries} experimental setting and each boxplot contains 200 points, one for the best run of the respective method on each of the missing masks (10 for each dataset) of each dataset (20 total). (a) shows the results for PARAFAC2 while (b) the respective results for tPARAFAC2, where {\small$\lambda=10$} and {\small$\lambda_B=100$}.}
    \label{fig:comparison}
\end{figure}

Figure \ref{fig:random_missing} demonstrates the accuracy of each method in terms of recovering the ground truth factors. As the percentage of missing data increases, it gets increasingly difficult for methods to recover the underlying patterns; therefore, FMS values start dropping for all methods. We also observe that incorporating ridge regularization does not enhance the recovery quality. However, tPARAFAC2 captures the underlying patterns more accurately by leveraging the temporal smoothness imposed on the evolving factors. While the improvement compared to PARAFAC2 is small at lower amounts of missing data, we observe significant improvement at higher amounts of missing data, especially at {\small$75\%$} missing data. Similar to the experiments with different amounts of noise, this improvement is attributed to tPARAFAC2's (a) ability to capture slowly changing patterns and (b) stronger noise reduction. Note that there are outliers in tPARAFAC2 with {\small$\lambda_B=10000$} in the {\small$75\%$} missing data case. These are due to one of the patterns being almost completely missing, which results in the respective factors being ``over smoothened", and hence achieving lower FMS.

We also compare the two different ways of handling missing data using AO-ADMM when fitting PARAFAC2 and tPARAFAC2 models. In terms of accuracy, AO-ADMM (EM) and AO-ADMM (RW) demonstrate similar performance. In terms of computational time, as shown in Figure \ref{fig:comparison}, as the percentage of unobserved data increases, the problem's difficulty increases, requiring more iterations and longer execution times. Furthermore, we observe that the AO-ADMM (EM) approach for fitting the models requires more iterations to converge than AO-ADMM (RW). However, each iteration is computationally less expensive, making the total execution time less. The primary computational bottleneck for the AO-ADMM (RW) method occurs during the update of the evolving mode factors, specifically the inversion of the term {\small$\M{D}_k \M{A}\Tra diag(\M{W}(:,j)) \M{A} \M{D}_k + \rho_{B_{k,j}}\M{I}$} across all indices {\small$j$} and {\small$k$} in \eqref{rw-update-b}. Moreover, adding temporal regularization increases computation time for datasets with {\small$25\%$} and {\small$50\%$} missing data. For datasets with {\small$75\%$} missing data, temporal regularization slightly reduces the total execution time (compared to PARAFAC2).
Overall, AO-ADMM (RW) is computationally more intensive, making AO-ADMM (EM) the preferred choice.

Experiments with higher percentages of missing data, e.g., 90\% missing data, indicate that tPARAFAC2 consistently recovers the ground truth with higher accuracy and the improvement is even larger. Comparing the two approaches of incorporating missing data we notice similar results between EM and RW for tPARAFAC2. For PARAFAC2, however, we notice that RW completely fails to recover the ground truth under these conditions (90\% missing data), with a median FMS of less than 0.2 over 200 runs, unless ridge (or temporal smoothness) regularization is imposed. The poor performance can be attributed to its update rules, specifically \eqref{rw-update-a}, \eqref{rw-update-b}, and \eqref{rw-update-c}. These rules require a sufficient number of data points or at least some prior knowledge through regularization, both of which are lacking in this setting.

\subsubsection{Structured missing data} \label{sec:structured}Frequently, missing data has a specific structure, rather than being random as in the previous experimental setting. For example, relevant words of an author might be missing at certain time points or information about a particular word could be consistently missing for that author. Here, we assess the performance of methods in the presence of such structured missing data.

We generate {\small$20$} datasets with dimensions {\small$100\times80\times25$}, introduce noise with {\small$\eta={0.75}$} and then, for each dataset, apply {\small$10$} binary indicator masks with {\small$10\%$}, {\small$25\%$} and {\small$40\%$} of mode-2 and mode-3 fibers fully missing. We do not consider mode-1 fibers missing as PARAFAC2 is not able to handle this case\footnote{If each slice is reconstructed as {\small$\M{X}_k \approx \M{A} \M{D}_k \M{B}_k\Tra$} and {\small$\M{X}_k(:,j)$} is fully missing, {\small$\M{B}_k(:,j)$} can be arbitrarily chosen as long as it satisfies all constraints, whereas on the other missing fiber cases, cross-slice or cross-column information can be leveraged.}. We then fit PARAFAC2 to each dataset using both ALS and AO-ADMM. Comparisons are also made with PARAFAC2 models incorporating ridge regularization across all modes, and tPARAFAC2 models, which are computed using both EM and RW updates. As in the previous setups, {\small$30$} initializations are shared across methods for each mask and each dataset.

\begin{figure}[!t]
\begin{minipage}[b]{\linewidth}
\centering      \centerline{\includegraphics[width=\textwidth]{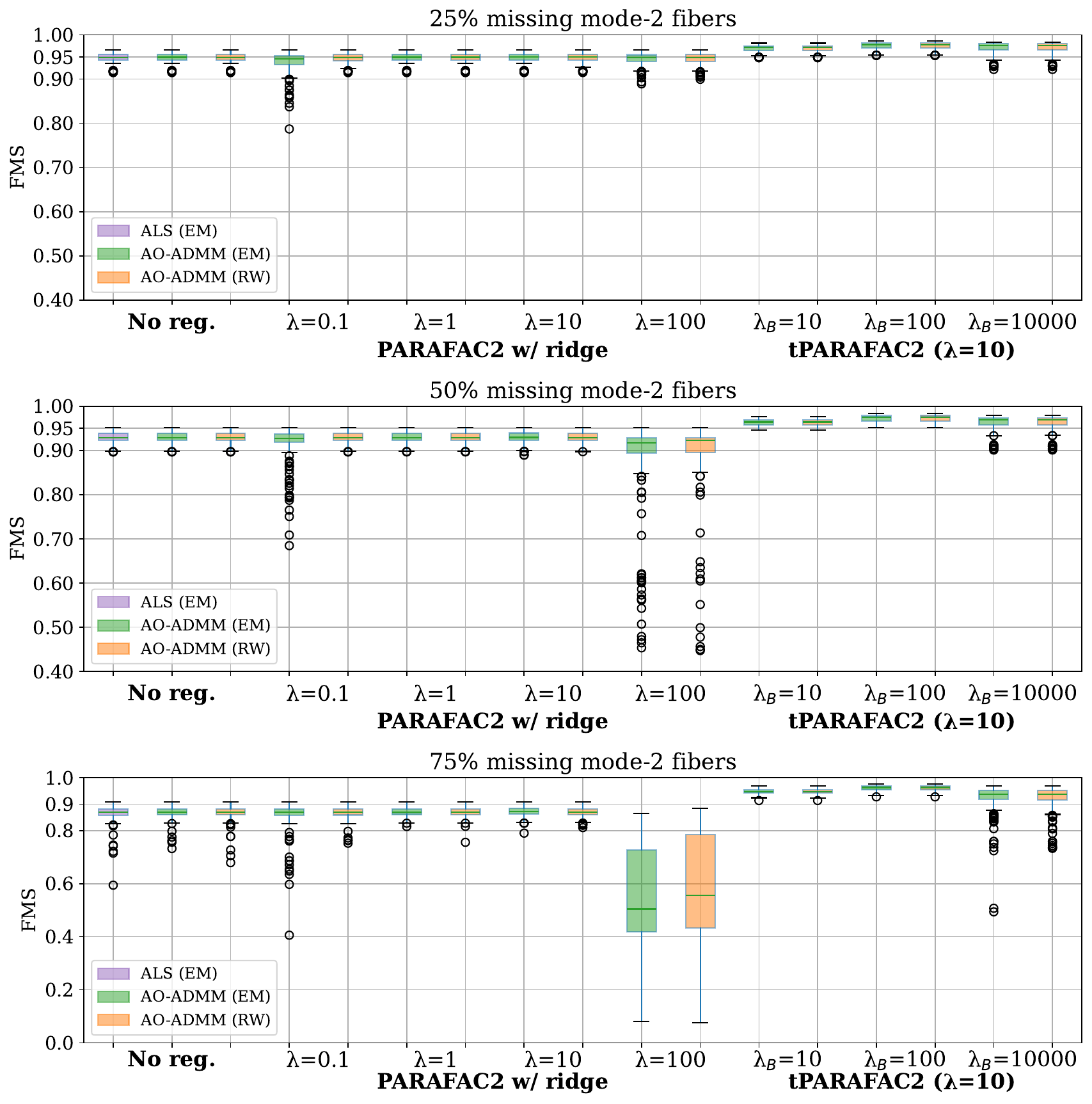}}
\caption{Accuracy of methods in terms of FMS with the ground truth, when mode-2 fibers are missing. Each boxplot contains {\small$200$} points, one for the best-performing run of the method at each mask of each dataset.}
\label{fig:fibers_2}
\end{minipage}
\end{figure}

\begin{figure}[t]
\begin{minipage}[b]{\linewidth}
\centering      \centerline{\includegraphics[width=\textwidth]{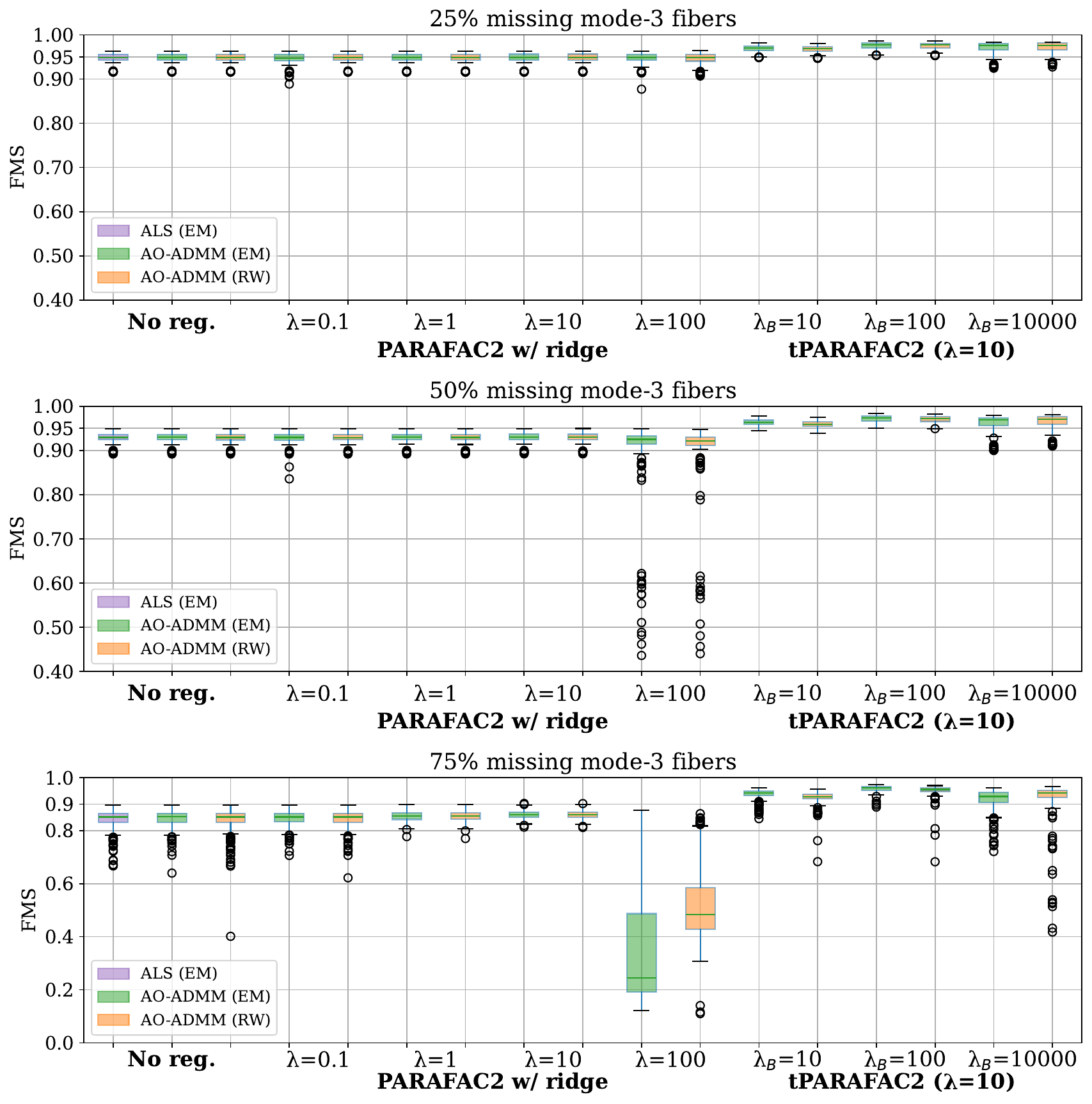}}
\caption{Accuracy of methods in terms of FMS with the ground truth, when mode-3 fibers are missing. Each boxplot contains {\small$200$} points, one for the best-performing run of the method at each mask of each dataset.}
\label{fig:fibers_3}
\end{minipage}
\end{figure}

Figures \ref{fig:fibers_2} and \ref{fig:fibers_3} show that, in general, as the amount of missing data increases, the quality of recovery for all methods decreases. tPARAFAC2 performs better compared to PARAFAC2 and PARAFAC2 with ridge regularization in all cases. Again, while the improvement is small at lower amounts of missing data, it becomes significant at higher levels of missing data. Accuracy-wise, the two approaches of handling missing data within the PARAFAC2 AO-ADMM framework perform similarly. Nevertheless, the RW approach, as also observed in the previous experiment, has higher computation times.

\subsection{Real data analysis}

\begin{figure}[t]
\begin{minipage}[b]{1.0\linewidth}
\centering      \centerline{\includegraphics[width=\textwidth]{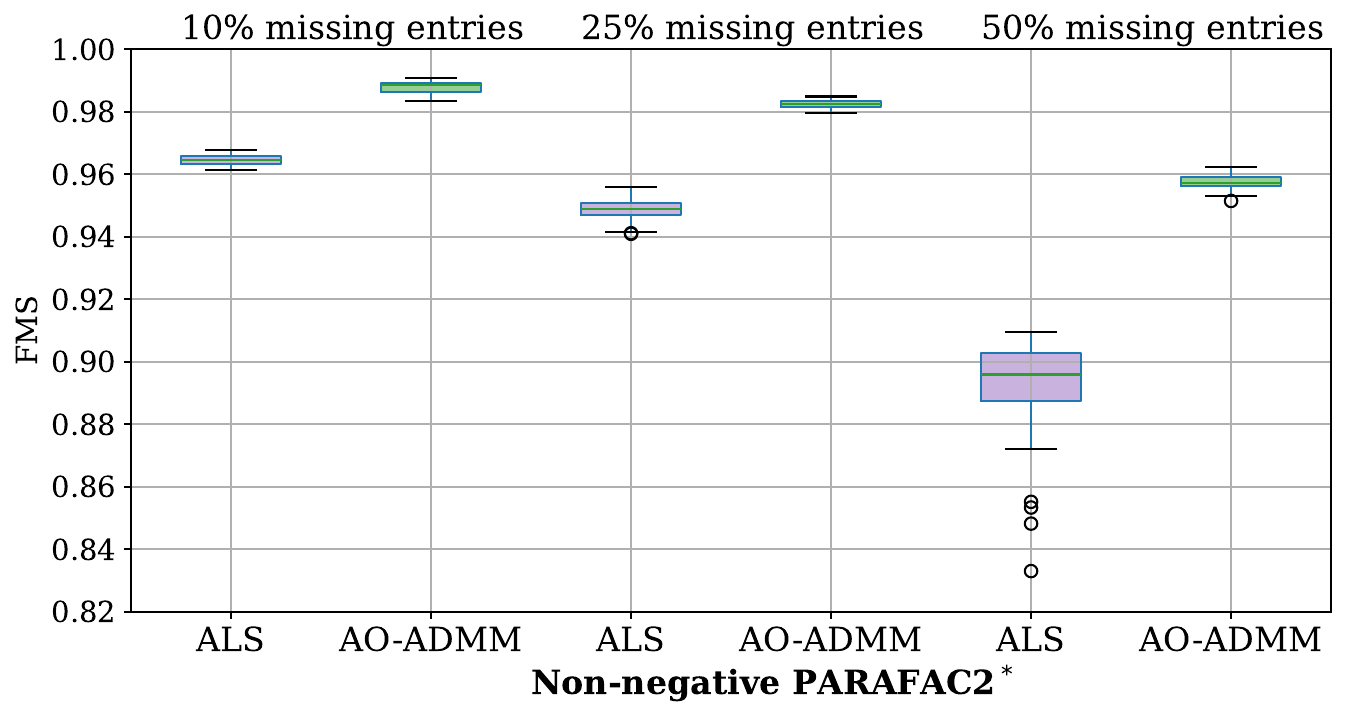}}
\caption{Chemometrics Application. Accuracy of extracted patterns from the GC-MS data using PARAFAC2 models measured by FMS with factors of fully observed data. Each boxplot contains 50 points, one for the best run of each method for each mask. $^{*}$: ALS has non-negativity constraints on the first and last modes, AO-ADMM on all three modes.}
\label{fig:gcms}\vspace{-0.25cm}
\end{minipage}
\end{figure}

\subsubsection{Chemometrics Application} In simulations, we have compared two different ways of handling missing data using the AO-ADMM algorithm for regularized PARAFAC2, and AO-ADMM(EM) has been the preferred choice. Here, we demonstrate on a chemometrics dataset that a PARAFAC2 model with constraints in all modes fitted using AO-ADMM(EM) to incomplete data can reveal the underlying patterns accurately. We use GC-MS (Gas Chromatography-Mass Spectrometry) measurements of apple wine samples, and analyze the data using PARAFAC2 with the goal of revealing the composition of mixtures (i.e., wine samples). The dataset is in the form of a third-order tensor with modes: 286 mass spectra, 95 retention time(s), and 57 wine samples.

PARAFAC2 has shown to be an effective approach to analyze such data and identify the compounds in the samples since the model allows for retention time shifts of the compounds in different samples \citep{bro1999parafac2}. Previously, PARAFAC2 with non-negativity constraints on all modes has been fitted to this data using AO-ADMM, demonstrating how constraints improve the interpretability of the model \citep{parafac2-aoadmm}.
Here, we demonstrate that such a model with constraints in all modes can be still used to analyze the data in the presence of missing data using AO-ADMM(EM). We show that constraints on the evolving mode improve the accuracy of the recovered patterns in the presence of missing data highlighting the importance of the extension of the AO-ADMM algorithm for regularized PARAFAC2 to incomplete data. The data is non-negative and we fit a 6-component PARAFAC2 model using AO-ADMM with non-negativity constraints on all modes. Each PARAFAC2 component reveals a mass spectrum ({\small${\M{A}}$}), elution profile ({\small $\{\M{B}_k\}_{k=1}^{57}$}) and the relative concentration of a chemical compound (modelled by this component) in the samples ({\small$\M{C}$}). In order to assess the performance in the presence of missing data, we randomly generate {\small$150$} binary indicator tensor masks ({\small$50$} for {\small$10\%$} missing data, {\small$50$} for {\small$25\%$} missing data, and {\small$50$} for {\small$50\%$} missing data) and compare the estimated patterns from data with missing entries using a PARAFAC2 model (fitted using ALS (EM) and AO-ADMM (EM)) with the patterns captured from the full data. When fitting PARAFAC2 using ALS (EM), we are only able to impose non-negativity on the first and third modes, while AO-ADMM imposes non-negativity constraints on all modes. For each mask, {\small $50$} initializations are randomly generated that are shared between methods and we only consider the solution using the initialization that results in the lowest function value (Equation \eqref{parafac2-aoadmm}).

\begin{figure*}[t]
\begin{minipage}[b]{1.0\linewidth}
\centering
\centerline{\includegraphics[width=1.0\textwidth]{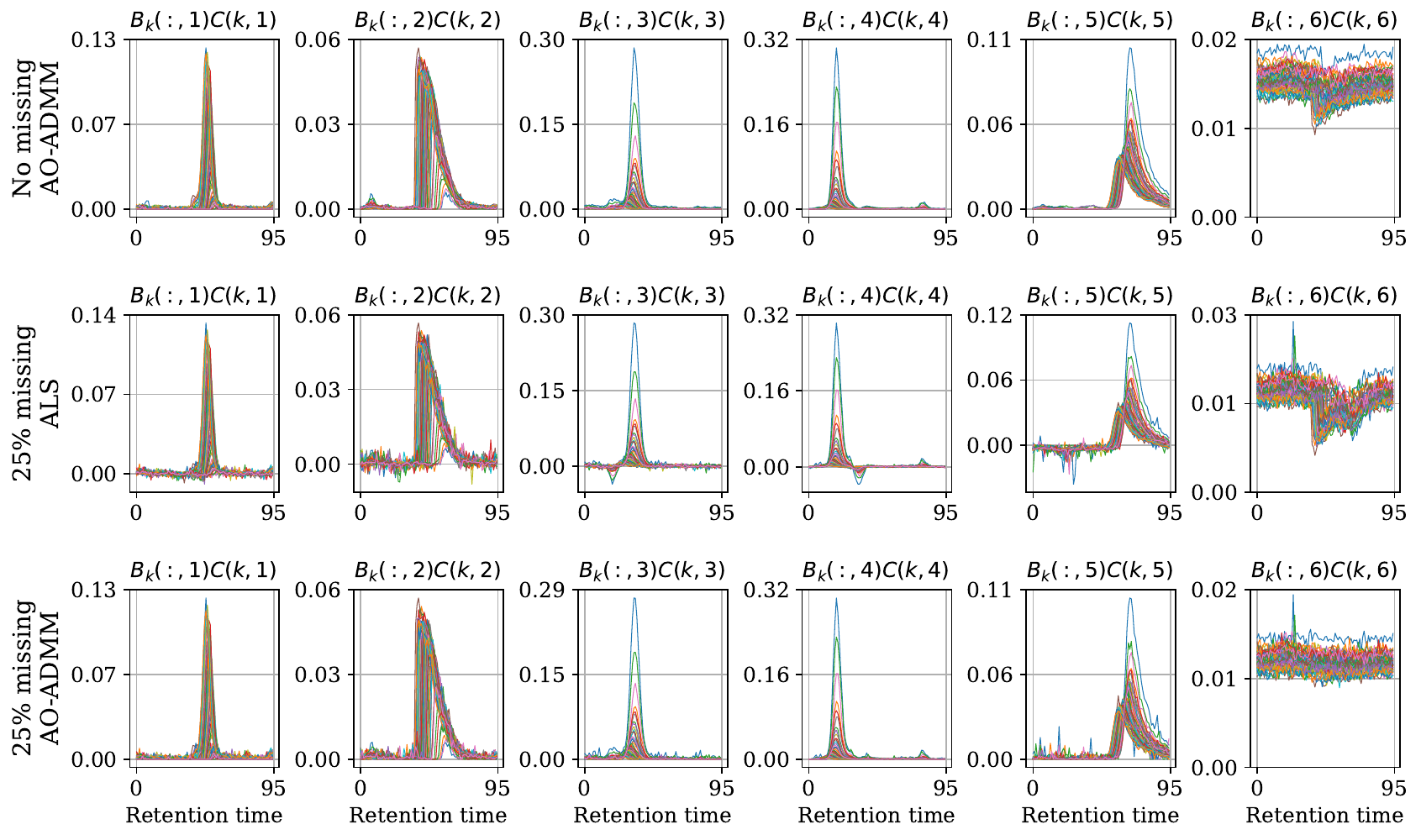}}
\caption{Chemometrics Application. Recovered factors from the retention time mode scaled by sample coefficient when no data is missing (first row), when {\small $25\%$} of the input is missing and ALS with non-negativity on the first and third mode is used to fit PARAFAC2 (second row), when {\small $25\%$} of the input is missing and AO-ADMM with non-negativity on all modes is used to fit PARAFAC2 (third row). Each colored line corresponds to different values of {\small$k$} (different samples).}
\label{fig:gcms_improvement}
\end{minipage}
\end{figure*}

Figure \ref{fig:gcms} shows the FMS values comparing the estimated factors from data with missing entries with the factors captured from the full data. We observe that incorporating the prior knowledge of the non-negativity of the factor improves the quality of the recovery. Thus, the AO-ADMM approach performs better. While both approaches achieve comparable accuracy in recovering the mass spectra {\small$\M{A}$} and sample concentration {\small$\M{C}$} factors, AO-ADMM demonstrates a distinct advantage in accurately recovering {\small$\{\M{B}_k\}_{k=1}^{57}$} corresponding to elution profiles. Figure \ref{fig:gcms_improvement} shows the uncovered profiles scaled by their concentration when no missing entries exist, alongside the factors when {\small$25\%$} of the input is missing (one of the {\small$50$} masks). The profiles obtained via ALS exhibit negative peaks, which are not realistic and hinder interpretation. In contrast, the non-negativity constraint applied in the AO-ADMM approach enhances accuracy by eliminating unrealistic negative components. For the full data, we note that ALS has an FMS of 0.98 with the solution given by AO-ADMM.

\subsubsection{Metabolomics Application}

In this application, we analyze dynamic metabolomics data to reveal evolving patterns in the metabolite mode. We also demonstrate the effect of temporal smoothness in the case of missing data. The dataset used in this experiment corresponds to Nuclear Magnetic Resonance (NMR) spectroscopy measurements and hormone measurements of blood samples collected during a meal challenge test from the COPSAC{\small$_{2000}$} (Copenhagen Prospective Studies on Asthma in Childhood) cohort \citep{copsac}. Blood samples were collected from participants after overnight fasting and at regular intervals after the meal intake (15 min, 30 min, 1 hr, 1.5 hr, 2 hr, 2.5 hr and 4 hr). To investigate metabolic differences among subjects in response to a meal challenge, these measurements have previously been analyzed using a CP model, revealing biomarkers of a BMI (body mass index)-related phenotype as well as gender differences \citep{yan2024characterizing}. Here, we use the measurements from males arranged as a third-order tensor with modes: 140 subjects, 161 metabolites and 7 time points. See \cite{yan2024characterizing} for more details about sample collection and data preprocessing.

\begin{figure}[t]
    \begin{minipage}[b]{1.0\linewidth}
      \centering      \centerline{\includegraphics[width=\textwidth]{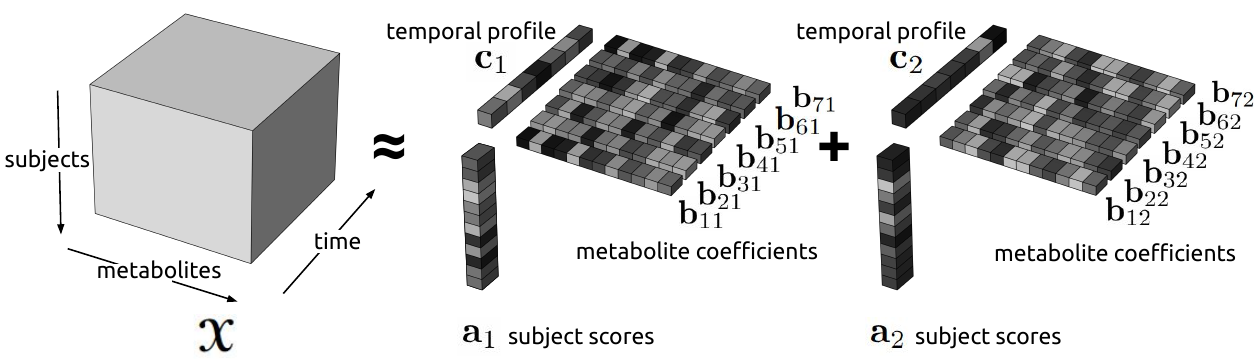}}
      \caption{2-component PARAFAC2 model of the metabolomics data with modes: subjects, metabolites, and time.}
    \label{fig:parafac2_nmr}
    \end{minipage}
\end{figure}

We analyze the data using 2-component PARAFAC2 and tPARAFAC2 models as in Figure \ref{fig:parafac2_nmr}. The number of components is selected based on the replicability of the components across subsets of subjects \citep{yan2024characterizing, reproducibility} (see supplementary material for more details). Also, a 2-component tCMF model is used as another baseline. While the CP model reveals the same metabolite mode factor for all time slices, PARAFAC2 allows metabolite mode factors to change in time. The component revealing BMI-related group difference is shown in Figure \ref{fig:nmr_factors}. In the supplementary material, we also compare CP, PARAFAC2, PARAFAC2 with ridge on all modes and tPARAFAC2 in terms of correlations with other meta-variables of interest (in addition to BMI). All in all, PARAFAC2 and tPARAFAC2 reveal evolving metabolite patterns while keeping the correlations comparable to CP. An animation of evolving PARAFAC2 factors is given \href{https://github.com/cchatzis/tPARAFAC2-latest/blob/main/animated_plot.gif}{in the GitHub repo}. Such evolving patterns reveal insights about the underlying mechanisms, e.g., the model captures the positive association between higher BMI and specific aminoacids, glycolysis-related metabolites, insulin and c-peptide at early time points - which disappears at later time points. 

\begin{figure}[t]
\begin{minipage}[b]{1.0\linewidth}
\centering
\centerline{\includegraphics[width=1.0\textwidth]{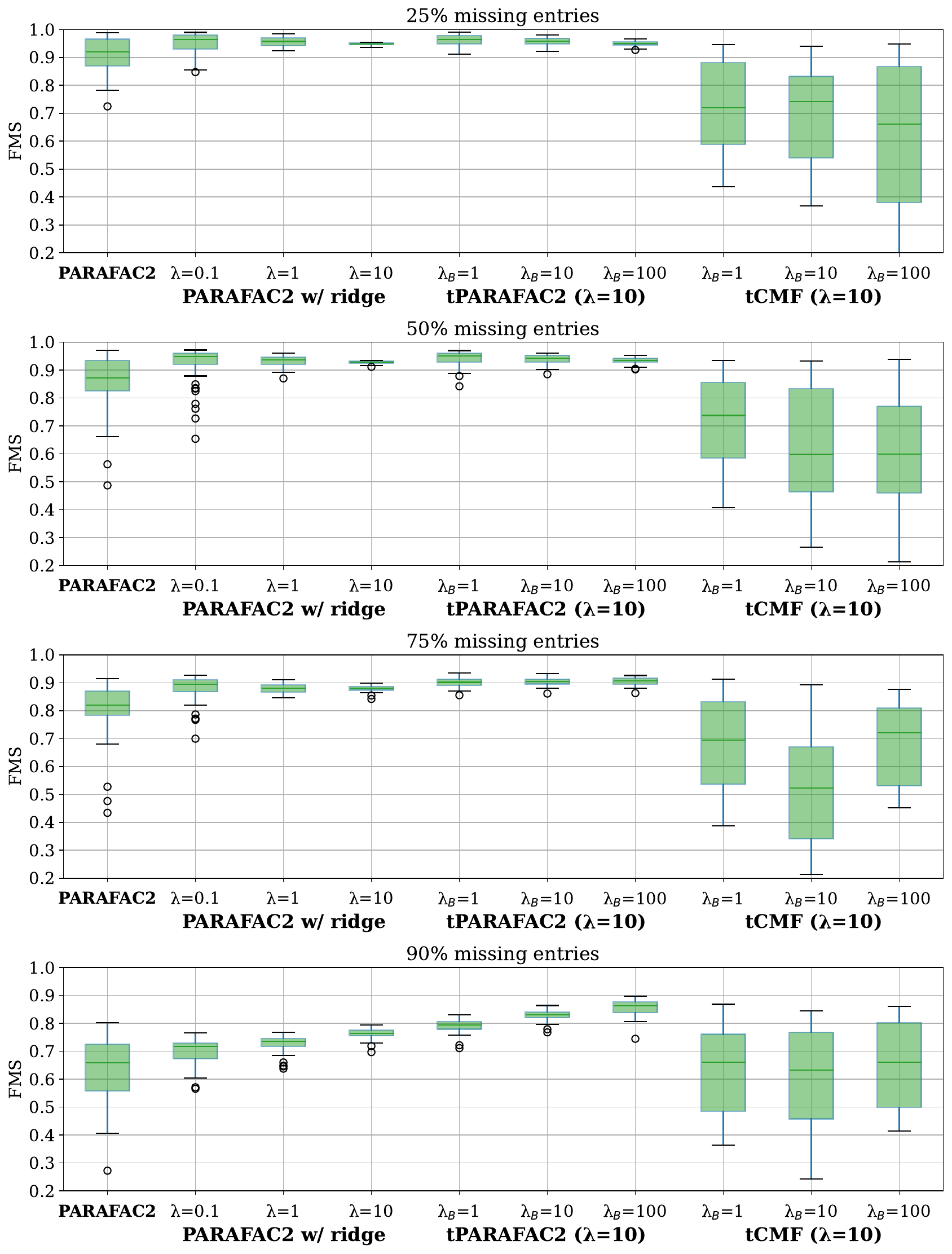}}
\caption{Metabolomics Application. FMS is computed using the factors extracted from the full dataset. Each boxplot contains 50 points, one for each random mask generated for each percentage.}
\label{fig:nmr}
\end{minipage}
\end{figure}

We also assess the performance of different models in the presence of missing data. Different amounts of missing data ({\small$25\%$}, {\small$50\%$}, {\small$75\%$} and {\small$90\%$}) are introduced, and we attempt to recover the patterns captured from the fully observed data using a PARAFAC2 model (fitted with AO-ADMM). For each percentage of missing data, we randomly generate {\small$50$} different indicator masks {\small$\T{W}$}, and for each mask {\small$50$} different random initializations are used (shared across all methods). The best run for each method and each mask is selected based on the lowest function value across all initializations. When generating each mask, we make sure that no full slices (any mode) or any mode-1 fibers are missing, and additionally, we do not allow for mode-3 fibers to be missing since missing measurements for a single metabolite for a single subject across time is not realistic. Instead, half of the missing entries in this experiment originate from fully missing mode-2 fibers (e.g.,  all measurements for a single subject at a single time point are missing), while the rest are randomly missing entries. We compare PARAFAC2, PARAFAC2 with ridge on all modes, tCMF and tPARAFAC2, where AO-ADMM is used to fit the models. Figure \ref{fig:nmr} shows that both ridge and temporal smoothness improve the recovery of the patterns, compared to PARAFAC2 without regularization. The tCMF model cannot capture the underlying patterns due to lack of uniqueness properties of the model. We observe that  PARAFAC2 with ridge and tPARAFAC2 have comparable performance for lower amounts of missing data. For $90\%$ missing data, however, tPARAFAC2 shows a clear improvement. We should note that such high amounts of missing data are not realistic in these measurements. Nevertheless, results are consistent with our experiments on simulated data, i.e., tPARAFAC2 enhances the recovery of underlying slowly changing patterns in temporal data, particularly for high levels of missingness.

\begin{figure*}[t]
\begin{minipage}[b]{1.0\linewidth}
\centering
\centerline{\includegraphics[width=1.0\textwidth]{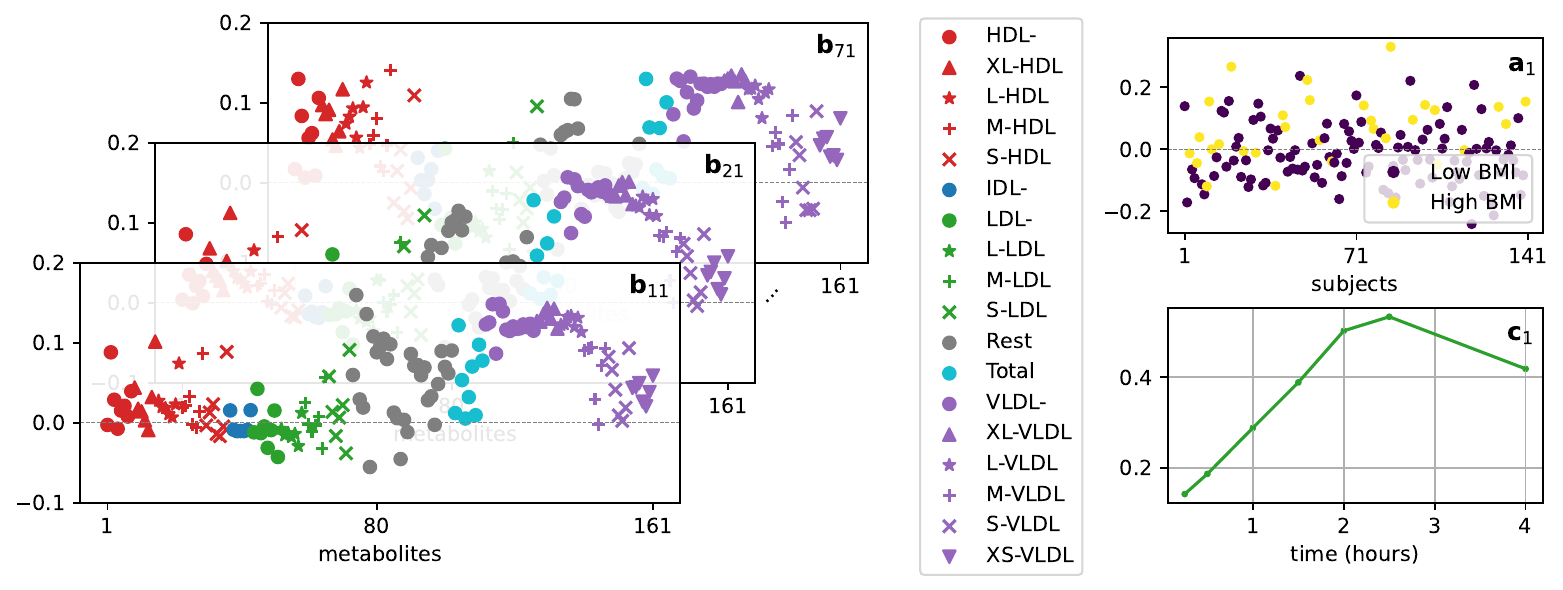}}
\caption{Metabolomics Application. The component showing statistically significant BMI group difference extracted from dynamic metabolomics data using PARAFAC2 (with non-negativity on the third mode). {\small $\V{b}_{k1}$}, for {\small $k=1,...,7$} corresponds to the factors in the metabolite mode changing in time. {\small $\V{a}_{1}$} shows the subject coefficients where subjects are colored according to BMI groups, and {\small $\V{c}_{1}$} shows the temporal profile. An animation of the evolving factors {\small $\V{b}_{k1}$} can be found \href{https://github.com/cchatzis/tPARAFAC2-latest/blob/main/animated_plot.gif}{alongside our code}.}
\label{fig:nmr_factors}
\end{minipage}
\end{figure*}

\section{Conclusion} \label{conclusion}
In this paper, we have introduced tPARAFAC2, a time-aware extension of the PARAFAC2 model, that incorporates temporal regularization with the goal of revealing the underlying evolving patterns in temporal data. We fit the model using an AO-ADMM-based algorithm. As temporal data is frequently incompletely observed, we have introduced two approaches for handling missing data within the PARAFAC2 AO-ADMM framework. Numerical experiments on both synthetic and real datasets demonstrate tPARAFAC2 can reveal the underlying evolving patterns more accurately compared to alternative approaches in the presence of high amounts of noise and missing data when the underlying patterns are slowly changing. We also demonstrate that among the considered missing data handling approaches, EM-based extension of AO-ADMM for regularized PARAFAC2 is both accurate and computationally more efficient.

One of the main limitations of the tPARAFAC2 approach is to determine the right parameter for temporal smoothness regularization. While in some experiments we have observed that the model performance is not that sensitive to the selection of the parameter, in some experiments, e.g., experiments with high amounts of noise, the model performance and the performance improvement depend on determining the right parameter for temporal smoothness regularization. In our experiments we have explored a parameter range up to the order of the norm of the data. It is also possible to scale the data by its Frobenius norm and consider parameters in much smaller scales. However, that is only a matter of scaling and selection of the parameters remains an open problem. Furthermore, we have only considered a single $\lambda_B$ value but it is possible that different patterns may have different levels of smoothness requiring different $\lambda_B$ values.
We plan to study this issue as future work to come up with a systematic parameter selection approach relying on replicability of the extracted patterns \citep{reproducibility}.

Another limitation of PARAFAC2-based approaches is the PARAFAC2 constraint. 
While the PARAFAC2 constraint has been crucial for uniquely capturing evolving patterns, real temporal datasets do not necessarily follow the PARAFAC2 constraint. However, a linear dynamical system-based regularization, temporal smoothness and the PARAFAC2 constraint are closely related \citep{ChSc25} and we plan to exploit that link further to develop time-aware factorization methods that can not only reveal the underlying evolving patterns uniquely but also forecast the patterns in future time steps. We also plan to study whether the constraint can be omitted and still reveal unique patterns when temporal data is jointly analyzed with other data sets \citep{ScCoAc21} inheriting uniqueness from other models such as the CP model.

\section*{Acknowledgements}

We would like to thank Rasmus Bro for providing the GC-MS data and insightful comments. This work was supported by the Research Council of Norway through project 300489 and benefited from the Experimental Infrastructure for Exploration of Exascale Computing (eX3) under contract 270053.

\bibliography{sn-bibliography}

\end{document}